\documentclass[11pt]{article}
\usepackage{authblk}

\usepackage[final]{acl}
\usepackage{booktabs}
\usepackage{longtable}
\usepackage{amsmath}
\usepackage{amssymb}
\usepackage{times}
\usepackage{latexsym}
\usepackage{multirow}
\usepackage{enumitem}
\usepackage{pifont}
\newcommand{\cmark}{\ding{51}} 
\newcommand{\xmark}{\ding{55}} 

\usepackage{float}
\usepackage{siunitx,threeparttable}
\usepackage[justification=raggedright,singlelinecheck=false]{caption}

\usepackage{microtype}
\usepackage{textgreek}
\usepackage{inconsolata}

\usepackage{graphicx}
\usepackage{natbib}
\usepackage{subcaption}

\usepackage[T1]{fontenc}
\usepackage[utf8]{inputenc}

\title{The \textit{Visual Iconicity Challenge}: Evaluating Vision--Language Models on Sign Language Form--Meaning Mapping}

\author{
\textbf{Onur Kele\c{s}}\textsuperscript{1,2},
\textbf{Asl{\i} {\"O}zy{\"u}rek}\textsuperscript{1,3},
\textbf{Gerardo Ortega}\textsuperscript{4},
\textbf{Kadir G{\"o}kg{\"o}z}\textsuperscript{2},
\textbf{Esam Ghaleb}\textsuperscript{1,3}
\\
\textsuperscript{1}Multimodal Language Department, Max Planck Institute for Psycholinguistics \\
\textsuperscript{2}Department of Linguistics, Bo\u{g}azi\c{c}i University \\
\textsuperscript{3}Donders Institute for Brain, Cognition and Behaviour, Radboud University \\
\textsuperscript{4}Department of Linguistics and Communication, University of Birmingham \\
\small \textbf{Correspondence:} \href{mailto:onur.keles1@bogazici.edu.tr}{onur.keles1@bogazici.edu.tr} and \href{mailto:esam.ghaleb@mpi.nl}{esam.ghaleb@mpi.nl}
}

\begin{document}
\maketitle
\begin{abstract}
Iconicity, the resemblance between linguistic form and meaning, is pervasive in sign languages, offering a natural testbed for visual grounding in vision–language models (VLMs). We introduce the \textit{Visual Iconicity Challenge}, a video-based benchmark that adapts psycholinguistic measures to evaluate VLMs on three tasks: (i) phonological sign-form prediction, (ii) transparency (inferring meaning from visual form), and (iii) graded iconicity ratings. We assess $17$ state-of-the-art VLMs in zero- and few-shot settings on Sign Language of the Netherlands and compare them to human baselines. 
VLMs mirror human phonological difficulty patterns (e.g., handshape harder than location) and achieve moderate to strong alignment with human iconicity ratings. However, most of them still fail to infer lexical meaning from visual form alone and show a systematic object-based bias that inverts the human preference for action-based signs. 
Crucially, \textit{models with stronger phonological form prediction correlate better with human iconicity judgments}, indicating shared sensitivity to visually grounded structure. Our findings validate these diagnostic tasks, show that explicit reasoning narrows the open-to-closed-model calibration gap, and motivate human-centric signals for modelling iconicity in multimodal models.
\end{abstract}

\begin{figure}
 \centering
  \includegraphics[width=0.8\linewidth]{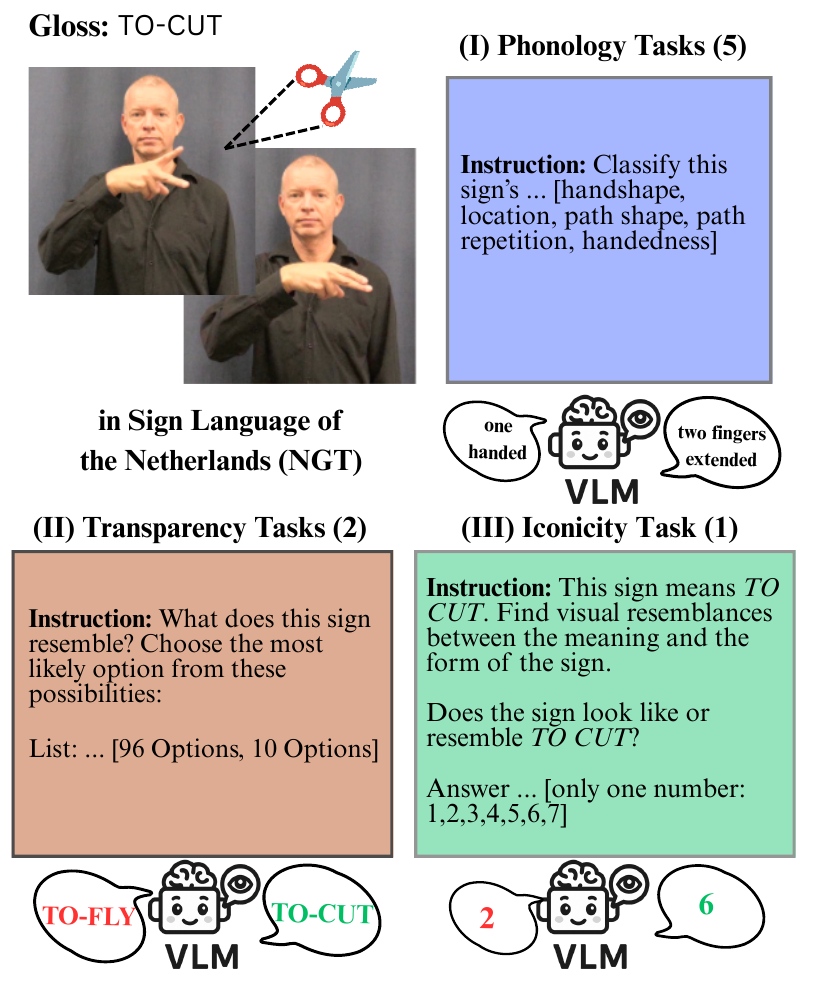}
  \caption{Overview of the \textit{Visual Iconicity Challenge}: evaluation pipeline of the sign \textsc{to-cut} in NGT for phonological form prediction, (top right), transparency (bottom left), and iconicity (bottom right) tasks. 
  }
  \label{fig:summary_figure}
\end{figure}

\section{Introduction}
Language is inherently multimodal: besides speech and text, it includes co-speech gesture and sign languages. Across these modalities, iconicity is the non-arbitrary link between form and meaning. Iconicity can be visual (e.g., speakers use iconic gestures through drawing shapes or trajectories in the air, adding depictive content alongside speech) or even vocal, as in onomatopoeia like “knock knock”, showing that form can transparently reflect meaning \cite{perlman2018people}. Within sign languages, iconicity is widespread. Estimates suggest that at least a third of lexical signs are iconic \cite{boyesbraem1986, campbell2025iconicity} and that between 50--60\% of signs' structure can be directly linked to the physical features of their referents \cite{ortega2017, pietrandrea2002}. They depict actions or shapes, providing a natural laboratory for studying the symbol grounding problem: how concepts connect to the physical world \cite{campbell2025iconicity, taub2001language}.

For vision–language models (VLMs), sensitivity to form–meaning mapping is a core test of grounding in human-centric signals \cite{bisk2020experience}. This is especially relevant for applications in sign language understanding and translation, as well as gesture and action recognition. A capable VLM should attend to \emph{dynamic} bodily movements and hand configurations---not just static objects or text---when interpreting a sign or gesture. sign languages offer a natural testbed for developing and evaluating models that must perceive temporally extended, simultaneous, visuospatial structure, rather than relying on static context alone \citep{yin2021including}. However, modern VLMs may exhibit static biases, over-attending to appearance features and under-attending to motion cues \citep{yu2025learning}, and relying on textual or contextual priors rather than visual evidence when processing gestures \citep{nishida-etal-2025-multimodal}.
To address these questions, we build the \textbf{Visual Iconicity Challenge}\footnote{The name is inspired by the “vocal iconicity challenge” of \citet{perlman2018people}.}: a sign dataset of Sign Language of the Netherlands (NGT), manually annotated with ground-truth phonological features, iconicity types, and iconicity ratings by non-signers, based on \citet{ortega_hearing_2019}. The dataset distinguishes between \emph{iconic} signs (with clear visual links to meaning) and \emph{arbitrary} signs (with no visual resemblance). 

We evaluate whether models capture different layers of sign–meaning structure, introducing three complementary tasks. 
Because iconicity links visual form to meaning, it depends on both \emph{phonological form competence} and \emph{analogical reasoning}, requiring models to map structured movement onto conceptual meaning through perceptuo-motor analogy \cite{thompson2019defining}. 
First, we test whether models can recognise the \emph{\textbf{phonological form of signs}}, including handshape, location, and movement features. 
Second, we examine \emph{\textbf{transparency}}: if a model can infer a sign’s intended \emph{meaning} from visual form alone \cite{hoemann1975transparency}, as non-signers often do \cite{sehyr2019perceived}.
Finally, we test their sensitivity to \emph{\textbf{iconicity}} itself, i.e., whether they can approximate human judgments of graded iconicity. Because previous work shows that VLMs may sometimes rely more on textual or contextual cues than on visual evidence \cite{nishida-etal-2025-multimodal}, the first two tasks also serve as checks that the models genuinely attend to the visual signal of the sign. Figure~\ref{fig:summary_figure} shows an overview of the three components. In summary, our contributions are: 
\begin{itemize}\setlength\itemsep{0pt}
    \item introducing the \emph{Visual Iconicity Challenge}, a benchmark of NGT signs with ground-truth sign phonological annotations and iconicity ratings; 
    \item collecting human baselines for phonology and transparency from a deaf signer and hearing sign-naïve participants;
    \item conducting the first large-scale zero- and few-shot assessment of state-of-the-art VLMs on sign language iconicity, analysing models' biases for object-based iconicity and failures of form–meaning transparency;
     \item releasing evaluation code, annotations, and human baselines via a repository for reproducibility and reuse.\footnote{\href{https://github.com/kelesonur/Visual_Iconicity_Challenge/}{https://github.com/kelesonur/Visual\_Iconicity\_Challenge}}

\end{itemize}



\section{Related Work}
\paragraph{Iconicity in language and computational models.} 
Iconicity has long been analysed as structure mapping between form and meaning in sign languages (e.g., depiction of shape or action) \cite{taub2001language, ortega2017, pietrandrea2002}. Psycholinguistic and lexical studies report substantial iconicity in signed lexicons and roles for iconicity in acquisition, processing, and L2 learning \cite{boyesbraem1986, campbell2025iconicity, karadoller2024iconicity, caselli2020degree}. 

Recent NLP research has explored analogous patterns in spoken language models. 
Large language models can capture sound symbolism effects.
For example, GPT-4 can generate iconic pseudowords whose meanings humans and models guess above chance \citep{marklova2025iconicity}. Furthermore, larger language models align with human iconicity ratings, indicating some sensitivity to sound symbolism \citep{loakman2024ears}. Metaphor understanding, like iconicity, depends on analogical mapping between domains \cite{lakoff1980metaphorical}.
\citet{tong2024metaphor} introduce the \textit{Metaphor Understanding Challenge}, which tests whether LLMs can interpret metaphors by distinguishing target-domain paraphrases from literal source-domain alternatives.  Their findings show that even advanced models often rely on surface similarity rather than analogy. 

In the visual modality, sound symbolism studies report weak or dataset-driven effects in CLIP/Stable Diffusion \citep{alper2023kiki} and mixed evidence for shape/magnitude symbolism in VLMs \citep{loakman2024ears}. 
Extending this understanding to the visual-manual modality of sign languages requires VLMs, which we address in this work. 

\paragraph{General multimodal benchmarks.} 
Large-scale multimodal benchmarks have assessed VLM capabilities on image captioning, VQA, and social signals. For example, \citet{zhang2025mmla} introduce MMLA, a suite of 61K multimodal utterances with labels for intent, emotion, style, etc., and report that even fine-tuned state-of-the-art models plateau around 60–70\% accuracy. Furthermore, \citet{li-etal-2025-multimodal-causal} introduce a Multimodal Causal Reasoning benchmark testing whether multimodal models can infer causal relations when crucial evidence appears in visual details. Their results show that models with strong textual reasoning still struggle with visual–conceptual integration. These resources and findings evaluate general multimodal capabilities but do not measure whether models map signed visual form onto meaning or assess graded iconicity relative to human judgments. 

\paragraph{Gesture and sign understanding with VLMs.} VLMs underperform on indexical/iconic gestures, especially with visuals-only input, indicating reliance on textual priors \citep{nishida-etal-2025-multimodal}. Systems like GIRAF mitigate this by injecting structured descriptors (pose skeletons, segmentations, depth) before LLM reasoning, achieving 75\% on deictic and 50\% on iconic gestures \citep{lin2023gesture}. Similarly, \citet{zhang2024pevl} introduce Pose-enhanced VLM, which integrates a skeletal pose modality into a CLIP-like model: one module uses the 2D pose to guide the visual attention to body joints, and another enriches the pose representation with visual context. This integration yields fine-grained action recognition by encouraging the model to focus on human motion cues. In sign language specifically, recent systems fuse additional signals. For example, SignLLM leverages human poses to generate sign language poses for digital human or avatar generation \cite{fang2024signllm}.

To our knowledge, no prior work has systematically probed VLMs on iconicity in sign languages. Our study is the first to do so at scale, evaluating how well off-the-shelf VLMs perceive the form–meaning transparency that signers exploit.

\section{Dataset: The Visual Iconicity Challenge}
We present a dataset built on the Sign Language of the Netherlands (NGT) from \citet{karadoller2024iconicity} and \citet{ortega_hearing_2019}.
It contains 96 sign videos (64 iconic signs and 32 arbitrary signs), each with an English gloss (meaning) and human iconicity ratings ranging from 1 to 7 (see the full dataset in Appendix \ref{stimuli}). This categorisation is based on human iconicity ratings: signs with low ratings (\textit{M} $=$ 2.10, \textit{SD} $=$ 0.50) were classed as arbitrary, and signs with high ratings (\textit{M} $=$ 5.13, \textit{SD} $=$ 1.02) were classed as iconic.\footnote{The original stimulus set distinguishes two iconic subgroups (high vs.\ low gesture-overlap with normed silent gestures), but as their iconicity ratings do not differ significantly \citep{karadoller2024iconicity}, we treat them as a single iconic category throughout.  The categories in the stimulus set were originally determined by consulting a native NGT signer and using a normed silent-gesture database \citep{ortega2020normed}. Also, throughout the paper, \textit{M} denotes the mean and \textit{SD} the standard deviation.} 

Although the benchmark is intentionally compact, the 96 signs were carefully balanced for iconicity, lexical class (verbs vs.\ nouns), and reference type (object/action/combined/arbitrary), and each sign carries dense psycholinguistic annotations (five phonological parameters, iconicity ratings, and human transparency baselines). This stratified design allows fine-grained per-feature and per-type analyses that would be infeasible at larger scale, and aligns with the size of psycholinguistic stimulus sets in the sign-language literature \citep{karadoller2024iconicity,ortega_hearing_2019}.
Our evaluation operationalises iconicity through three complementary tasks targeting different form--meaning mapping levels. 
\emph{(i) \textbf{Phonological form prediction}} examines whether models perceive the \emph{articulatory structure} of a sign (handshape, location, movement). \emph{(ii) \textbf{Transparency}} asks models to recover a sign’s \emph{lexical meaning} from visual form alone, indexing analogical mapping from form to concept while minimising reliance on linguistic priors. \emph{(iii) \textbf{Graded iconicity rating}} evaluates whether models are sensitive to the \emph{degree} of resemblance between form and meaning by correlating model ratings with human judgments.

\noindent\textbf{Hypothesis.} Models that better predict phonological features (e.g., handshape, location, path) should better capture iconicity, since both require grounding in structured bodily properties.

\begin{table}[!htbp]
  \centering
  \footnotesize
  \setlength{\tabcolsep}{4pt}
  \renewcommand{\arraystretch}{0.95}
  \begin{tabular}{@{}p{0.45\linewidth}cc@{}}
  \toprule
  Item & Ortega et al. & Ours \\
  \midrule
  Phonology form features {\tiny (based on Klomp and Pfau \citeyear{KlompPfau2020_NGT})} & &  \\
  \quad Handshape &  \xmark & \cmark \\
  \quad Location &  \xmark &  \cmark\\
  \quad Path shape &  \xmark &  \cmark\\
  \quad Path repetition &  \xmark & \cmark \\
  \quad Handedness &  \xmark & \cmark \\
  \addlinespace[6pt]
  Transparency & & \\
  \quad labels (N=96) & \cmark & \cmark \\ 
  \addlinespace[6pt]
  Iconicity & & \\
  \quad Ratings (1–7) & \cmark & \cmark \\ 
  \quad Labels (\tiny{Iconic vs.\ arbitrary}) & \cmark & \cmark \\ 
  \quad Types (\tiny{e.g., Object or action based}) & \xmark & \cmark \\\addlinespace[2pt]
  \addlinespace[6pt]
  
  \midrule
  Human baselines &  &  \\
  \quad Phon. form prediction & \xmark & \cmark  \\
  \quad Transparency & \xmark & \cmark  \\
  \quad Iconicity ratings & \cmark & \cmark \\
  \bottomrule
  \end{tabular}
  \caption{Comparison of the original NGT sign videos dataset \citep{ortega_hearing_2019, karadoller2024iconicity} and our extensions for the visual iconicity challenge.}
  \label{tab:dataset_comparison}
  \end{table}

Motivated by the view that iconicity relies on both \emph{phonological form competence} and \emph{analogical reasoning} \cite{thompson2019defining}, we extend the original resource with:  
(i) detailed phonological annotations for each sign,  
(ii) iconicity-type labels, and  
(iii) human baselines for phonology and transparency.  
These additions support evaluation of VLMs from sub-lexical perception to graded iconicity. See Table \ref{tab:dataset_comparison} for a comparison between the original dataset and our extensions.

\subsection{Sign Phonological Form Features} 
\label{sect:phonological_features}
\begin{figure}
  \centering
  \includegraphics[width=0.9\linewidth]{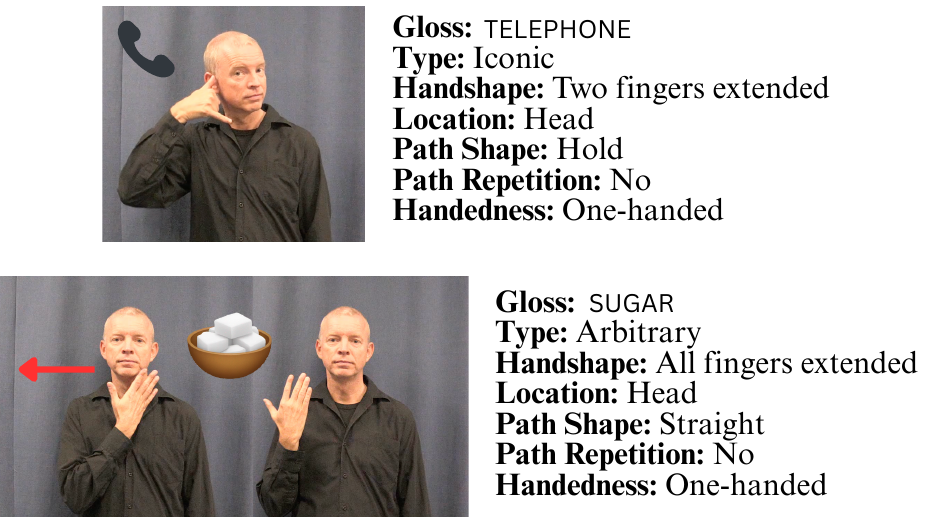}
  \caption{Examples of an iconic vs. an arbitrary sign, with their annotated phonological form features. The sign \textsc{telephone} is iconic as its form resembles a telephone’s shape, whereas \textsc{sugar} is arbitrary with no clear visual link to its meaning.}
  \label{fig:handshape_examples}
\end{figure}

We annotate phonological form features of each sign using a standard NGT phonology framework~\cite{KlompPfau2020_NGT}. These are discrete, visual descriptors of articulation (elaboration on the annotation criteria is in Appendix~\ref{criteria}). In summary, we use five phonological parameters:

\begin{itemize}
    \item \textit{Handshape}: 7 categories (e.g., fist, flat hand, one finger extended, etc.). Figure \ref{fig:handshape_examples} illustrates a few categories of the annotated handshapes.
    \item \textit{Location}: 5 categories of where on the signer’s body or space the sign is articulated, i.e., face/head, torso, arm/shoulder, the opposite hand, or neutral space.
    \item \textit{Path Shape}: 4 categories of movement trajectory shape, i.e., no movement/hold, straight line, arched curve, and circular motion.
    \item \textit{Path Repetition}: 2 categories (whether the movement is repeated or only single).
    \item \textit{Handedness}: 3 categories (one-handed sign, two-handed symmetrical, or two-handed asymmetrical). 
\end{itemize}

A deaf signer and a hearing non-signing researcher performed the annotations. To assess reliability, inter-annotator agreement ranged from 77.9\% ($\kappa=0.73$) for handshape to 98.9\% ($\kappa=0.98$) across parameters. All disagreements were discussed and resolved.  
These reliable annotations serve as a gold-standard reference or “ceiling” for assessing how well models can recognise sign form.

\textbf{Task.} Given a sign video, models perform multi-class prediction for each parameter: handshape, location, path shape, path repetition, and handedness (prompts in Appendix~\ref{prompts}). For this task, we report the accuracy of the model predictions per parameter and the overall average accuracy. This task checks whether the model is capable of extracting form information from the video: where and how signs were articulated.

\textbf{Human baseline.} We gather baseline results from human participants for all the phonological parameters.
The 96 stimulus signs were divided into four lists of 24 signs each. Four sign-naïve Turkish undergraduate participants (i.e., without prior knowledge of sign language) were recruited and compensated with course credit, and randomly assigned to lists in a counterbalanced design. Each participant judged all 24 signs in their list on both the phonological feature tasks and the transparency (open-set meaning identification) task. This provided a sign-naïve baseline for both tasks. The human baseline mean phonological accuracy was 0.79 (highest for handedness, lowest for handshape).

\subsection{Sign Transparency}
\label{sec:transparency}
\paragraph{Task.} Transparency tests whether meaning can be inferred from visual form alone. We use the gloss list (i.e., meaning) of each sign, which is provided in the original dataset. We evaluate two settings: \textbf{Transparency\textsubscript{1}} (open-set identification among all 96 glosses) and \textbf{Transparency\textsubscript{2}} (multiple choice with 10 candidates: the target gloss plus 9 distractors). We use accuracy as the primary metric (proportion of signs correctly identified).

\textbf{Human baseline.} The deaf signer (who annotated the phonological form features and is not a native NGT signer) identified $57/96$ glosses in the open-set setting; the sign-naïve group identified $40/96$ (same participants and lists as in the phonology baseline). These provide upper- and lower-bound human references for Transparency\textsubscript{1}.

\subsection{Sign Iconicity Ratings}
\label{sect:iconicity}
\paragraph{Task.} This task probes whether models capture the degree to which a sign’s form resembles its meaning. 
We use the original crowdsourced iconicity ratings as \textbf{human baselines }(see Appendix~\ref{stimuli}).
 Each sign has an average iconicity rating on a 1–7 scale (with 7 = “looks exactly like its meaning”, 1 = “not iconic at all”). 
Models are prompted to produce the iconicity rating for each sign (i.e., the degree of the sign's resemblance to its meaning).
We report Spearman's $\rho$ (rank alignment) and quadratic-weighted Cohen's $\kappa_w$ (ordinal scale agreement) between model ratings and the average human iconicity ratings.

\paragraph{Iconicity types.} Iconicity type influences how signers perceive, process, and acquire signs \cite{ortega2014type, ortega2017type}. We annotate each sign for its iconicity type to probe how well models align with these distinctions. These include \emph{object-based} signs ($N=16$), where the handshape visually resembles a property of the referent (e.g., the wings of a butterfly), and \emph{action-based} signs ($N=31$), where the hand depicts an action performed on or by the referent (e.g., brushing teeth). The remaining 17 signs belonged to a third category named ``combined'', where both strategies were employed for the same sign.
The descriptions of these types can be found in Appendix \ref{iconicity_examples}.
These label types enable us to analyse how different iconic strategies affect model predictions and human perception.

\begin{figure*}
  \centering
  \includegraphics[width=1\linewidth]{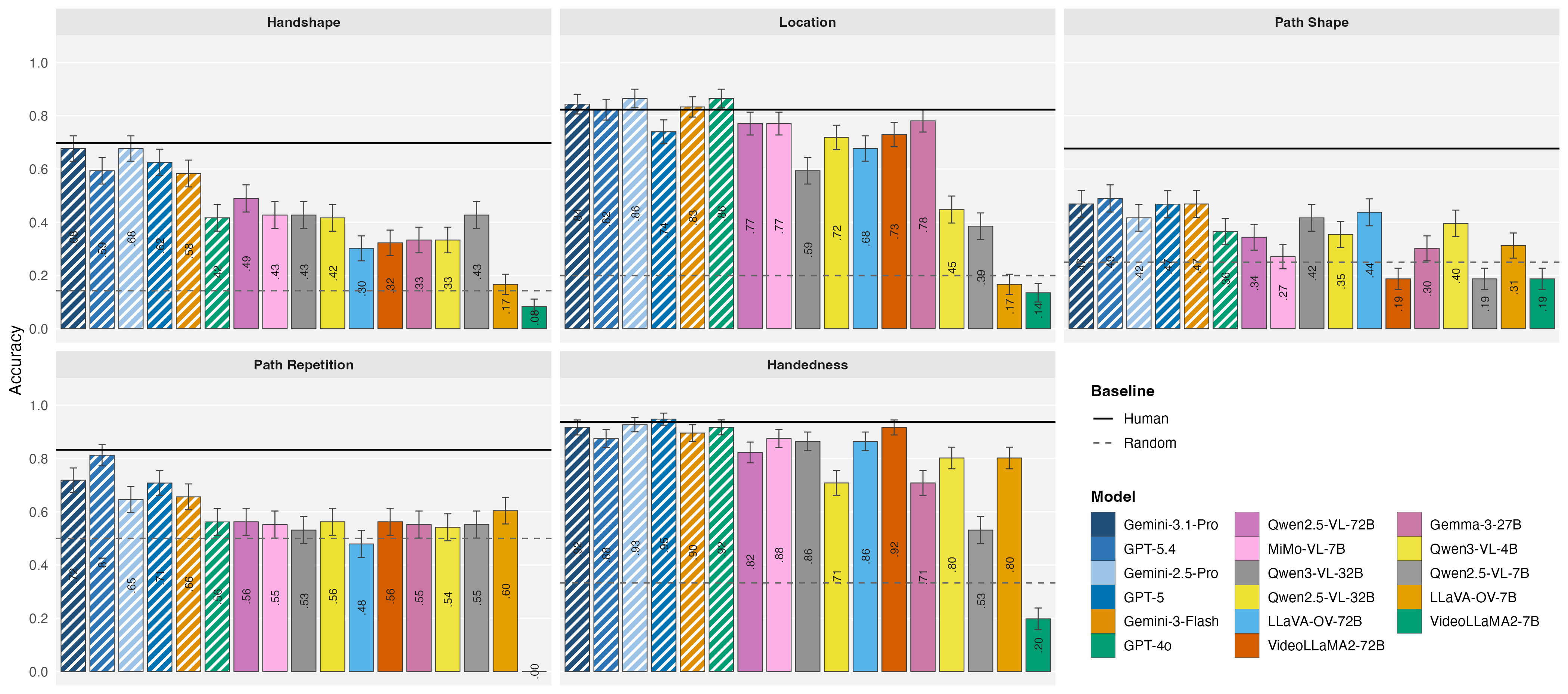}
  \caption{Zero-shot accuracy per phonological feature for all 17 VLMs. Striped bars: closed-source; solid bars: open-source. Solid black line: deaf-native human baseline; dashed grey line: random baseline. Error bars are binomial SEs over 96 signs. Numbers inside bars are mean accuracies.}
  \label{fig:all_form_feature_results}
\end{figure*}

\section{Models and Inference}
\paragraph{Models.}
We evaluate a representative and diverse set of 11 open-source VLMs and 6 proprietary models, spanning small ($\leq$7B), medium (27--32B), and large ($\geq$70B) parameter regimes. Open models: Qwen2.5-VL (72B/32B/7B) \cite{bai2025qwen2}, Qwen3-VL (32B/4B), VideoLLaMA2 (72B/7B) \cite{videollama2_2024}, LLaVA-Onevision (72B/7B) \cite{li2024llava}, Gemma-3 (27B) \cite{team2025gemma}, and MiMo-VL-7B \cite{mimo2025}. Proprietary models: GPT-4o \citep{openai2024gpt4o}, GPT-5 \cite{openai2025gpt5systemcard}, GPT-5.4 \cite{openai2026gpt54systemcard}, Gemini~2.5~Pro \cite{comanici2025gemini25}, Gemini~3~Flash \cite{googledeepmind2025gemini3flash}, and Gemini~3.1~Pro \cite{googledeepmind2026gemini31pro}.
For inference, smaller models ($\leq$7B) were run on a single NVIDIA A100 GPU, while larger models ($\geq$27B) were distributed across up to four A100 GPUs. Closed-source models were queried via API calls.

\paragraph{Technical details.}
Our dataset comprises 96 sign language videos, all recorded at 50 fps, with mean duration of $2.42 \pm 0.40$ seconds (range: 1.68--3.24\,s). All models sample 8 frames per video and process each through their native vision encoder. As a concrete example, Qwen3-VL-32B uses dynamic-resolution encoding, yielding $\sim$1{,}165 vision tokens per video (8 frames). In the 4-shot configuration, each shot contains video ($\sim$1{,}165 tokens) + question ($\sim$86 tokens) + answer ($\sim$1 token) = $\sim$1{,}252 tokens, so the total input is 4 shots (5{,}008 tokens) + test (1{,}251 tokens) = 6{,}259 tokens, representing 2.4\% of the 262{,}144-token context window. Other models use similar architectures with varying vision encoders.

\paragraph{Zero-shot setup.}
All models are evaluated in a zero-shot manner first. We craft a prompt/instruction template for each task that is standardised across models. The prompts explicitly describe the task and the expected answer format, and we ensure the output format is constrained (e.g., just a single number for ratings, or a one-word answer for glosses). For example, for the iconicity rating task, the prompt to the model is:

\begin{quote}\small
\emph{This sign means: <MEANING>. Some signs are iconic, and some are arbitrary. Find visual resemblances between the meaning and the form of the sign. How much does the sign look like ``<MEANING>''? Answer with only one number: 1,2,3,4,5,6,7 (1=not at all, 7=exactly).}
\end{quote}

We do not use chain-of-thought prompting or specialised prompting tools, as initial trials with those did not show clear benefits. Our aim is to first establish baseline performance; more sophisticated prompting or fine-tuning can be explored in future work. For Qwen3-VL-32B, we additionally test with thinking mode enabled (denoted~$^\dagger$) on the iconicity task, to probe whether explicit reasoning improves scale calibration.

\paragraph{Few-shot setup.}
To examine whether a few examples can improve models' performance, we conduct 4-shot experiments on a representative panel of open models that span the size spectrum: Qwen2.5-VL (72B/32B/7B), Qwen3-VL (32B/4B), Gemma-3-27B, and MiMo-VL-7B. We omit closed models in these settings since few-shot probing suggests that GPT-5 and Gemini variants are already comparatively well-calibrated in zero-shot settings, showing only marginal benefit.

We provide 4 example QA pairs (two iconic signs and two arbitrary signs) before the test query, using the same instruction format and showing the correct outputs for those examples.

\section{Results \& Discussions}
\label{sect:results}

\subsection{Phonological Form Prediction}
\paragraph{Zero-shot.}
Figure \ref{fig:all_form_feature_results} shows the accuracy of each model on the five phonological sub-tasks.
The location of the sign (where on the body the sign is articulated) and handedness (one vs. two hands) were the easiest.
For location, $11/17$ models reach $\geq0.70$ (best $0.865$), while for handedness, $15/17$ exceed $0.70$. In contrast, handshape and path shape remain the hardest features, with only closed-source models exceeding 0.50 on handshape. Most models exceed the random baseline but remain below the human mean of 0.794.
Among closed models, Gemini~3.1~Pro leads with mean accuracy \textit{M} $=$ 0.725, followed by GPT-5.4 (\textit{M} $=$ 0.719), both surpassing Gemini~2.5~Pro (\textit{M} $=$ 0.706); Gemini~3~Flash (\textit{M} $=$ 0.688) falls just below it. Among open-source models, Qwen2.5-VL-72B (\textit{M} $=$ 0.598), MiMo-VL-7B (\textit{M} $=$ 0.579), Qwen3-VL-32B (\textit{M} $=$ 0.567), and LLaVA-OV-72B and Qwen2.5-VL-32B (both \textit{M} $=$ 0.552) lead. Overall, while large models encode phonologically relevant structure, the absolute gap to human performance remains large for most features. Full per-feature results for all models are in Appendix~\ref{sect:full_results_phonology} (Table~\ref{tab:accuracy_by_model_handshape_sorted_mean}) and Appendix~\ref{sect:additional_results} (Figure~\ref{fig:form_features_summary}).

Interestingly, the performance patterns of models for each form feature in Figure \ref{fig:all_form_feature_results} mirror well-established acquisitional asymmetries in sign language. Like deaf children and adults \citep{kelecs2022effects,sandler2006sign,morgan2007phonology,MarentetteMayberry1999}, models find \emph{location} easier than \emph{handshape}.

\paragraph{Few-shot.}
Few-shot prompting yields modest, model-dependent gains for selected open-source VLMs (Table~\ref{tab:oneshot_vs_fourshot_acc}).
Qwen2.5-VL-7B shows the largest absolute gain ($0.417 \to 0.550$), followed by Qwen2.5-VL-32B and Qwen3-VL-32B, while Qwen2.5-VL-72B changes little. MiMo-VL-7B is essentially flat. This pattern indicates that few-shot prompting mainly benefits models with lower zero-shot performance, presumably by helping them interpret the task format.\footnote{A closer look at task-level breakdowns reveals uneven effects: the largest gains occur for path shape and handedness, while location remains unstable, handshape shows minimal improvement, and path repetition is largely unaffected. This suggests that few-shot prompting primarily helps models disambiguate structural features like path shape and handedness.}

\begin{table}[!htbp]
\small
\centering
\begin{tabular}{l@{\hspace{10pt}}cc}
\toprule
\multirow{2}{*}{Model} & \multicolumn{2}{c}{Mean Accuracy} \\
\cmidrule(lr){2-3}
& 0-shot & 4-shot \\
\midrule
Qwen2.5-VL-72B   & 0.598 & 0.600 \\
Qwen2.5-VL-32B   & 0.552 & 0.620 \\
Qwen3-VL-32B     & 0.567 & 0.633 \\
Gemma-3-27B      & 0.535 & 0.572 \\
MiMo-VL-7B    & 0.579 & 0.573 \\
Qwen2.5-VL-7B    & 0.417 & 0.550 \\
Qwen3-VL-4B      & 0.504 & 0.573 \\
\bottomrule
\end{tabular}
\caption{Comparison of zero-shot and 4-shot performance on the phonological form prediction task.}
\label{tab:oneshot_vs_fourshot_acc}
\end{table}


\subsection{Sign Transparency}
\paragraph{Zero-shot.}
Open-set gloss identification is highly challenging (Table~\ref{tab:transparency_acc}). Even the strongest closed-source models perform poorly: the best model, Gemini~3.1~Pro, identifies 28 of the 96 glosses ($\approx$29.2\%), followed by Gemini~3~Flash (19/96, 19.8\%) and Gemini~2.5~Pro (17/96, 17.7\%). All remain well below the human baselines (57/96 for the deaf signer of T\.{I}D evaluated cross-linguistically, 40/96 for hearing non-signers).\footnote{Importantly, the deaf signer is a native signer of Turkish Sign Language (T\.{I}D) and is \emph{not} familiar with NGT. Their score therefore reflects a Transparency task (inferring meaning from unfamiliar visual forms), not a lexical retrieval task, and serves as a cross-linguistic upper bound on how much iconicity supports inference of meaning across sign languages.}
Restricting the task to a 10-way multiple-choice format improves scores (e.g., 47/96 for Gemini~3.1~Pro, 44/96 for Gemini~3~Flash, 42/96 for GPT-5, and 41/96 each for Gemini~2.5~Pro and GPT-5.4).
Open-source VLMs perform substantially worse, with the best (Qwen3-VL-32B) achieving only 4/96 correct identifications in the open setting and 19/96 in the 10-way setting. The persistent advantage of closed models over all open-source systems indicates that high-capacity proprietary VLMs are better at leveraging combined visual and linguistic cues.

Across models, correct predictions cluster on visually obvious signs such as \textsc{telephone}, \textsc{pistol}, and \textsc{to-wring} (guessed by the large majority of evaluated VLMs), followed by \textsc{to-juggle}, \textsc{windscreen wiper}, and \textsc{deer} (see Figure~\ref{fig:transparency_common_glosses} in Appendix~\ref{sect:additional_results}). Interestingly, some arbitrary but cross-linguistically shared signs (e.g., \textsc{to-argue}, \textsc{to-order}, \textsc{person}, and \textsc{to-die}) were successfully guessed by a sizable number of VLMs.\footnote{Most of these arbitrary signs were guessed correctly by our human participants too.}
Identification of such arbitrary signs suggests that their forms still contain strong visual cues, possibly through conventional metaphorical mappings shared across sign languages \citep{meir2018metaphor}.

\begin{table}[!htbp]
  \small
  \centering
  \begin{tabular}{lrr}
  \toprule
  Model & 96 options & 10 options \\
  \midrule
  \multicolumn{3}{l}{\textit{\textbf{Human baselines}}}\\
  Deaf signer (T\.{I}D)             & 0.594 & --   \\
  Hearing non-signer       & 0.417 & --   \\
  \addlinespace
  \multicolumn{3}{l}{\textit{\textbf{Models}}}\\
  Gemini-3.1-Pro          & \bfseries 0.292 & \bfseries 0.490 \\
  Gemini-3-Flash          & 0.198 & 0.458 \\
  GPT-5                   & 0.156 & 0.438 \\
  Gemini-2.5-Pro          & 0.177 & 0.427 \\
  GPT-5.4                 & 0.146 & 0.427 \\
  GPT-4o                  & 0.073 & 0.354 \\
  Qwen3-VL-32B            & 0.042 & 0.198 \\
  Qwen2.5-VL-32B          & 0.052 & 0.177 \\
  MiMo-VL-7B           & 0.042 & 0.177 \\
  Qwen2.5-VL-72B          & 0.021 & 0.167 \\
  Qwen3-VL-4B             & 0.021 & 0.167 \\
  LLaVA-OV-72B      & 0.031 & 0.156 \\
  VideoLLaMA2-72B         & 0.031 & 0.156 \\
  Gemma-3-27B             & 0.021 & 0.125 \\
  VideoLLaMA2-7B          & 0.010 & 0.125 \\
  Qwen2.5-VL-7B           & 0.021 & 0.115 \\
  LLaVA-OV-7B       & 0.021 & 0.073 \\
  \midrule
  Chance (random)         & 0.010 & 0.100 \\
  \bottomrule
  \end{tabular}
  \caption{Transparency task accuracy in 96-option and 10-option conditions. The deaf signer is a native signer of Turkish Sign Language (T\.{I}D), \emph{not} NGT.}
  \label{tab:transparency_acc}
  \end{table}
  
\paragraph{Few-shot.}
Four-shot prompting yields no meaningful gains. In the 96-way setting, Qwen2.5-VL (72B \& 32B) and Gemma-3-27B each identify 2 of 92 items, and Qwen2.5-VL-7B identifies 1 of 92. In the 10-choice format, the same models score 15–16 of 92, matching their zero-shot levels. These results suggest the bottleneck is not just understanding the task format, but a fundamental limitation in the models’ visual–semantic grounding.

\subsection{Sign Iconicity}
\paragraph{Zero-shot.}
For iconicity ratings (Table~\ref{tab:iconicity_results}), several models show moderate-to-strong positive correlations with human iconicity judgments ($\rho \geq 0.40$, $p<.001$). Gemini~3.1~Pro achieves the highest rank alignment with human ratings ($\rho = 0.774$), followed by GPT-5.4 ($\rho = 0.612$), GPT-5 ($\rho = 0.607$), and Gemini~3~Flash ($\rho = 0.600$). 
Among open VLMs, Qwen2.5-VL-72B shows the strongest correlation ($\rho = 0.501$), followed by Qwen3-VL-32B with thinking mode enabled ($\rho = 0.473$), Qwen2.5-VL-7B ($\rho = 0.456$), and Gemma-3-27B ($\rho = 0.452$). Open models such as Qwen2.5-VL-7B ($\kappa_w = 0.421$) and Gemma-3-27B ($\kappa_w = 0.459$) approach mid-tier closed-model calibration (e.g., Gemini~2.5~Pro: $\kappa_w = 0.458$), despite lower $\rho$, indicating better scale calibration relative to their rank alignment.

Importantly, model--human alignment is multifaceted: rank consistency ($\rho$) and ordinal agreement ($\kappa_w$) capture complementary aspects of relative ordering and scale calibration. Gemini~3.1~Pro leads on both dimensions ($\rho = 0.774$, $\kappa_w = 0.522$), with GPT-5 and GPT-5.4 following closely ($\kappa_w = 0.511$ and $0.499$, respectively). Several open models (e.g., Qwen2.5-VL-7B, Gemma-3-27B) match closed-model scale calibration ($\kappa_w$) despite weaker rank discrimination ($\rho$), suggesting they use the rating scale appropriately but are less sensitive to relative iconicity differences between signs, while a subset substantially under-rate iconicity (Qwen2.5-VL-32B, VideoLLaMA2-72B, MiMo-VL-7B, VideoLLaMA2-7B). Despite these strengths, most models still compress the scale around the midpoint and systematically over-rate arbitrary signs, thereby reducing the contrast between iconic and arbitrary categories.

Notably, thinking mode substantially benefits Qwen3-VL-32B ($\rho = 0.473$, $\kappa_w = 0.459$ vs.\ $\rho = 0.356$, $\kappa_w = 0.177$ without thinking), resolving its scale-anchoring failure and lifting it to second place among open models. Its $\kappa_w$ of $0.459$ is joint-highest among open models (tied with Gemma-3-27B) and matches closed-model levels (e.g., Gemini~2.5~Pro: $0.458$), suggesting that reasoning helps models adopt a more human-like scale, not just improve relative rankings.

\begin{table}[!htbp]
\centering
\small
\begin{threeparttable}
\begin{tabular}{l
                S[table-format=1.3] @{} l
                S[table-format=1.3]}
\toprule
Model & {$\rho$} & & {$\kappa_w$} \\
\midrule
Gemini-3.1-Pro          & \bfseries 0.774 & \textsuperscript{***} & \bfseries 0.522 \\
GPT-5.4                 & 0.612 & \textsuperscript{***} & 0.499 \\
GPT-5                   & 0.607 & \textsuperscript{***} & 0.511 \\
Gemini-3-Flash          & 0.600 & \textsuperscript{***} & 0.396 \\
Gemini-2.5-Pro          & 0.577 & \textsuperscript{***} & 0.458 \\
GPT-4o                  & 0.248 & \textsuperscript{*}   & 0.227 \\
\midrule
Qwen2.5-VL-72B          & 0.501 & \textsuperscript{***} & 0.284 \\
Qwen3-VL-32B$^\dagger$  & 0.473 & \textsuperscript{***} & 0.459 \\
Qwen2.5-VL-7B           & 0.456 & \textsuperscript{***} & 0.421 \\
Gemma-3-27B             & 0.452 & \textsuperscript{***} & 0.459 \\
VideoLLaMA2-72B         & 0.400 & \textsuperscript{***} & 0.261 \\
MiMo-VL-7B              & 0.389 & \textsuperscript{***} & 0.179 \\
Qwen3-VL-32B            & 0.356 & \textsuperscript{***} & 0.177 \\
Qwen2.5-VL-32B          & 0.344 & \textsuperscript{***} & 0.152 \\
Qwen3-VL-4B             & 0.238 & \textsuperscript{*}   & 0.122 \\
LLaVA-OV-72B            & 0.223 & \textsuperscript{*}   & 0.172 \\
LLaVA-OV-7B             & 0.119 & \textsuperscript{ns}  & 0.112 \\
VideoLLaMA2-7B          & 0.101 & \textsuperscript{ns}  & 0.047 \\
\bottomrule
\end{tabular}
\end{threeparttable}
\caption{Graded iconicity rating results. Significance codes for $\rho$: ${}^{*} p<.05$, ${}^{**} p<.01$, ${}^{***} p<.001$, ns $p\ge .05$. ${}^\dagger$Thinking mode enabled during inference.}
\label{tab:iconicity_results}
\end{table}

\paragraph{Few-shot.}
Few-shot prompting helps Qwen2.5-VL-32B ($\rho: 0.344 \to 0.510$) and Qwen3-VL-32B ($0.356 \to 0.488$) most, with smaller but consistent gains for Gemma-3-27B ($0.452 \to 0.484$) and MiMo-VL-7B ($0.389 \to 0.412$) (Table~\ref{tab:iconicity_zerovs4shot}). Few-shot cues are thus most helpful for models that underperform relative to their capacity.

\begin{table}[!htbp]
\small
\centering
\begin{tabular}{lcc}
\toprule
\multirow{2}{*}{Model} &
\multicolumn{2}{c}{Spearman $\rho$} \\
\cmidrule(lr){2-3}
& 0-shot & 4-shot \\
\midrule
Qwen2.5-VL-72B   & 0.501 & 0.521 \\
Qwen3-VL-32B     & 0.356 & 0.488 \\
Qwen2.5-VL-32B   & 0.344 & 0.510 \\
Gemma-3-27B      & 0.452 & 0.484 \\
MiMo-VL-7B       & 0.389 & 0.412 \\
\bottomrule
\end{tabular}
\caption{Comparison of zero-shot and 4-shot performance on the graded iconicity rating.}
\label{tab:iconicity_zerovs4shot}
\end{table}

\begin{figure}[t]
    \centering
    \includegraphics[width=1\linewidth]{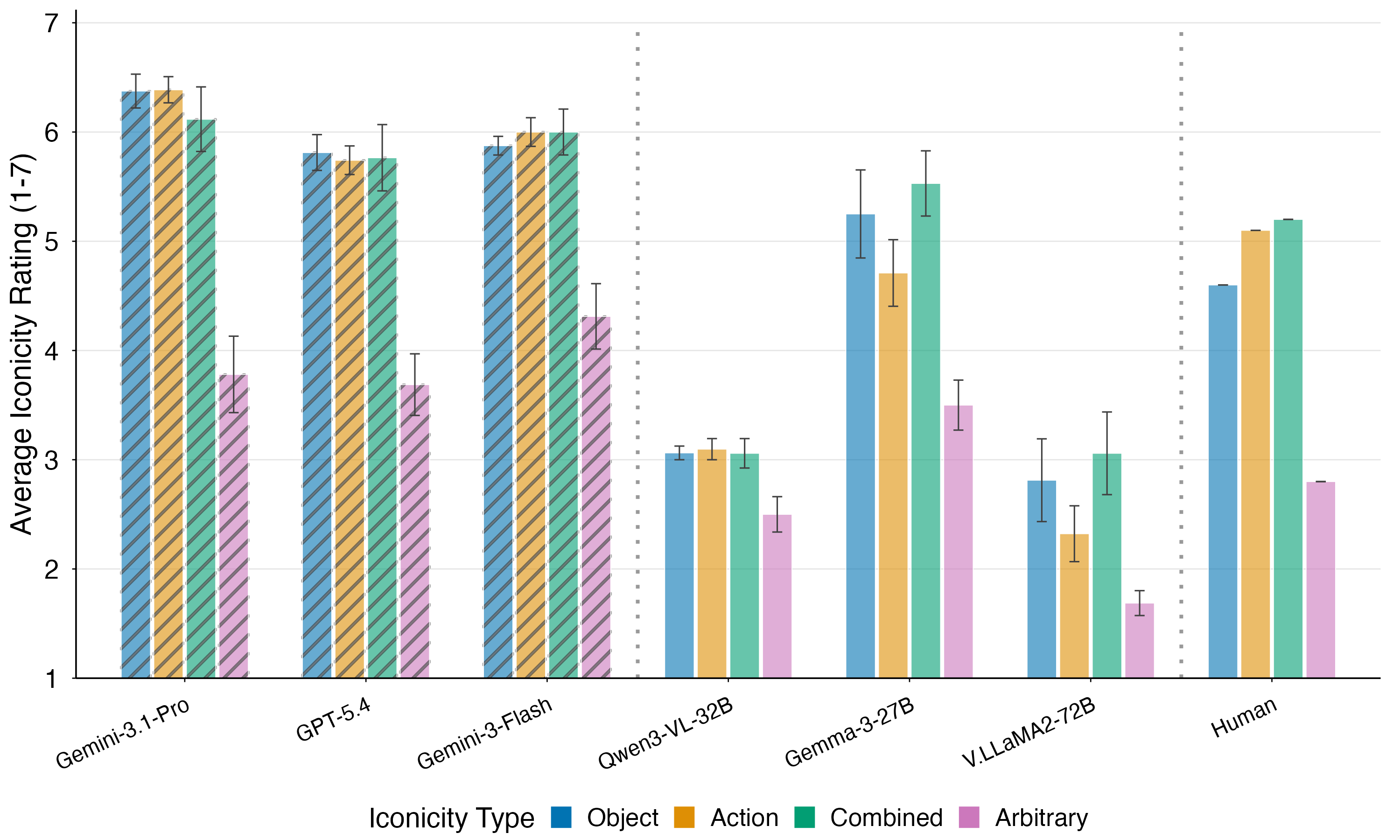}
    \caption{Average iconicity ratings by iconicity type for closed-source (striped) and open-source (solid) models, with the human baseline right of the dotted divider. Error bars: SE across signs. Colours denote reference type: \textcolor{blue}{Object}, \textcolor{orange}{Action}, \textcolor{green!50!black}{Combined}, \textcolor[RGB]{255,182,193}{Arbitrary}.}
    \label{fig:iconicity_type}
\end{figure}

\paragraph{Type of iconicity.}
Iconicity is commonly classified by whether a sign depicts an object's shape or a human action \citep{ortega_hearing_2019}. Human raters show a robust preference for \emph{action}-based signs over \emph{object}-based ones, consistent with findings that action signs are acquired earlier and that both deaf children and adults exhibit a cognitive bias toward action-based (handling) iconic forms \citep{ortega2017type,sumer2025action}. As illustrated in Figure~\ref{fig:iconicity_type}, both humans and large models clearly distinguished arbitrary from iconic signs, indicating that models can broadly recognise iconic structure. However, within iconic signs, differences emerged. Humans show a consistent \emph{action} bias, whereas most open-source models displayed the reverse pattern, favouring \emph{object}-based signs that depict visual features rather than actions. Closed-source models such as Gemini and GPT-5 showed little to no preference between the two types. This static-image bias is well known in computer vision \citep{Zhou_2025_ICCV}, but our contribution is exposing its \emph{linguistic} consequence: a systematic inversion of human iconicity preferences that links a known visual-modeling limitation to a new, linguistically meaningful failure mode (see our qualitative observations in Appendix~\ref{sect:error_analysis}).

\section{Interaction of Iconicity, Transparency, and Phonology}
\label{sect:interactions}
We hypothesise that models with stronger phonological form predictions are better at both rating graded iconicity and inferring sign meaning, as all three tasks require grounding in structured bodily properties.

\begin{figure}[th]
 \centering
  \includegraphics[width=1\linewidth]{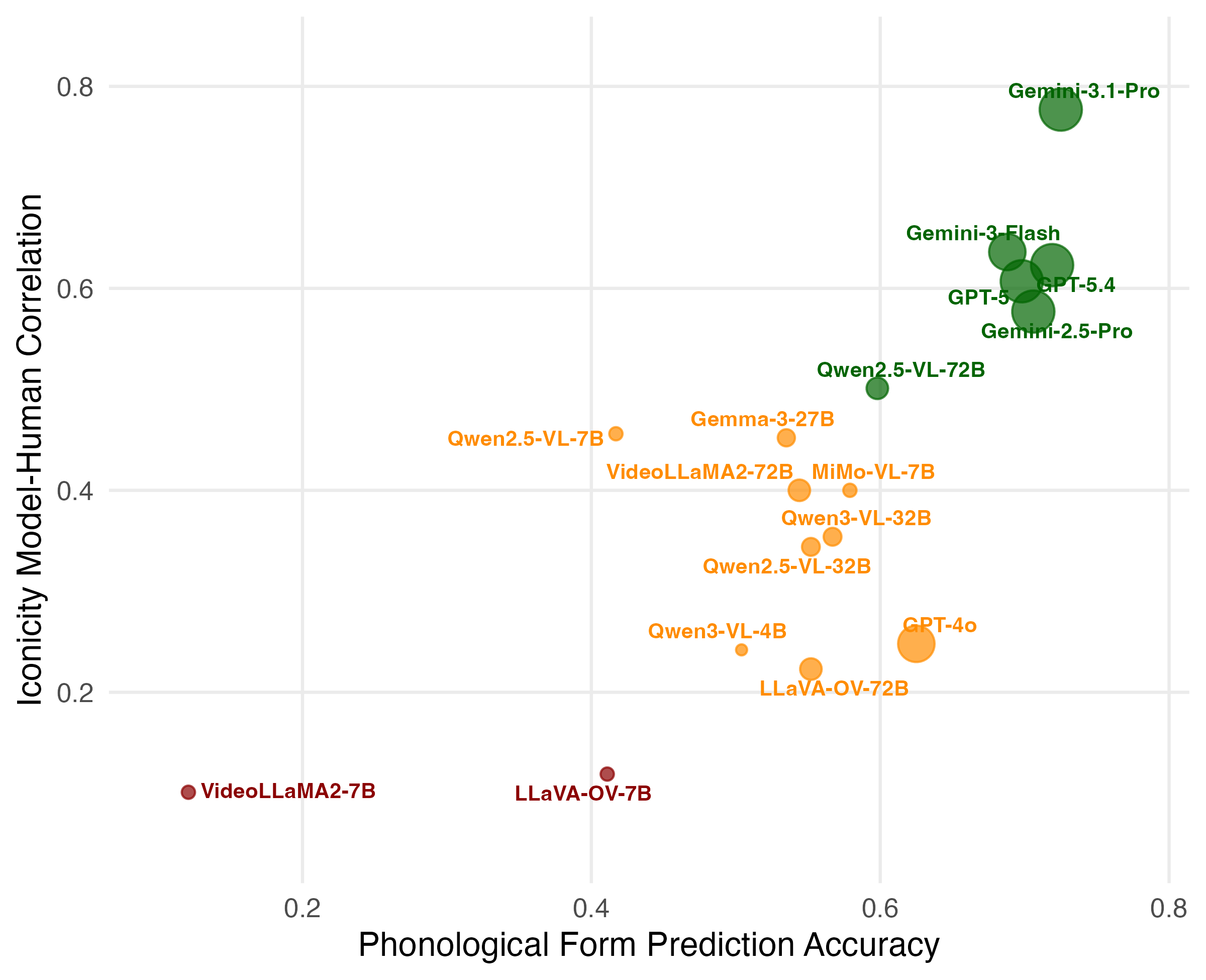}
    \caption{Overall model landscape by zero-shot phonological form prediction accuracy ($x$-axis) and iconicity Spearman $\rho$ with humans ($y$-axis). Top-right is best; dot size encodes model size. Colours denote performance tiers: \textcolor{red!70!black}{red} = low, \textcolor{orange!90!black}{orange} = mid, \textcolor{green!50!black}{green} = high (assigned by the average of normalised phonology and iconicity scores).}
  \label{fig:p5}
\end{figure}

\paragraph{Phonology $\leftrightarrow$ Iconicity.}
As shown in Figure~\ref{fig:p5}, models with higher phonological form accuracies, such as Gemini~3.1~Pro, Gemini~3~Flash, GPT-5.4, GPT-5, and Gemini~2.5~Pro, also achieve closer alignment with human iconicity ratings. Conversely, models with weaker phonological representations (e.g., smaller Qwen and LLaVA variants) show both lower accuracy on phonological features and less consistent treatment of iconicity. Across all 17 models, mean phonological accuracy correlates strongly with iconicity $\rho$ ($\rho = .769$, $p < .001$), suggesting that sensitivity to phonological form and to form--meaning mappings are not independent, but may sign language co-develop \cite{emmorey2014iconicity}.

\paragraph{Phonology $\leftrightarrow$ Transparency.}
Mean phonological accuracy correlates strongly with transparency accuracy (96-opt: $\rho = .898$; 10-opt: $\rho = .913$; both $p < .001$), indicating that explicit form encoding substantially supports meaning inference regardless of whether the decision space is constrained. Together with the phonology--iconicity link above, our three tasks operationalise complementary aspects of form--meaning mapping: phonological perception of bodily structure, analogical inference from form to concept (transparency), and graded sensitivity to resemblance (iconicity).

This link mirrors the human case, where iconicity is tied to phonological awareness because mapping form features onto conceptual structure requires attending to form and meaning simultaneously. Yet models systematically overrate object-based iconicity (Figure~\ref{fig:iconicity_type}), so phonological sensitivity alone does not yield the human action-based preference.

\section{Conclusion}
\label{eom:conclusion_start}
We introduced the \emph{Visual Iconicity Challenge}, a diagnostic evaluation that probes phonological form prediction, meaning prediction from form, and iconicity ratings in the Sign Language of the Netherlands. Our evaluations suggest that, compared to human baselines, larger vision large language models mirror human phonological difficulty patterns (e.g., handshape is more difficult than location), can distinguish iconic from arbitrary signs, and correlate moderately to strongly with graded human iconicity ratings. Yet they fail to infer lexical meaning and show a different iconicity-type bias. Bridging this gap requires richer gesture/sign pretraining and dynamic pose encoding; explicit reasoning offers a complementary route, as thinking mode alone lifted open-model scale calibration to closed-model levels.
\label{eom:end_of_main}

\section{Limitations}
Our evaluation has several constraints. The dataset is small (96 isolated NGT signs) with citation-style clips that may not generalise to other sign languages or continuous discourse. Phonological annotations cover five major parameters but omit finer-grained features (orientation, aperture changes, non-manual markers), and the mixed lexical classes (verbs vs. nouns) may affect transparency and iconicity patterns. 

Furthermore, we evaluated models only in zero-shot and few-shot settings without fine-tuning, which establishes a diagnostic baseline but likely underestimates potential performance with sign-specific training. Future work may explore fine-tuning, examine model-specific factors (parameter count, memory footprint, mixture-of-experts activation patterns during inference to examine the processing of signs with different levels of iconicity), test robustness under visual perturbations (noise, motion blur), conduct stratified analyses by iconicity type (action-based vs. object-based) and sign difficulty levels, and perform qualitative error analysis to identify whether failures stem from visual perception, analogical reasoning, or lexical-semantic grounding deficits.

\section*{Ethics Statement}
All human participants provided informed consent prior to participation. The study was approved by the Institutional Review Board in Social Sciences and Humanities at Boğaziçi University.

\small
\bibliography{custom}

\appendix

\renewcommand{\thesection}{\Alph{section}}  

\onecolumn
\section{Stimuli}
\label{stimuli}
\subsection*{Iconic Signs (n = 64)}
\begin{longtable}{lc|lc}
\toprule
\textbf{Sign} & \textbf{Iconicity rating} & \textbf{Sign} & \textbf{Iconicity rating} \\
\midrule
\endhead
TO-BREAK & 6.79 & TO-INJECT & 5.0 \\
TO-CRY & 6.74 & TO-STAPLE & 4.89 \\
WINDSCREEN WIPER & 6.63 & CALCULATOR & 4.8 \\
TO-CUT & 6.61 & PENGUIN & 4.78 \\
ELEPHANT & 6.53 & RATTLE & 4.75 \\
BICYCLE & 6.44 & CAR & 4.7 \\
BIRD & 6.42 & CURTAINS & 4.7 \\
BABY & 6.39 & BRIDGE & 4.6 \\
KEY & 6.26 & DEER & 4.6 \\
TELEPHONE & 6.22 & HELICOPTER & 4.6 \\
TO-WRING & 6.12 & MONKEY & 4.5 \\
TO-SWIM & 6.11 & SPIDER & 4.5 \\
TO-SLAP & 6.11 & ZIMMER & 4.42 \\
TO-PUMP & 6.11 & TO-ERASE & 4.2 \\
PIANO & 6.05 & TO-SMS & 4.2 \\
TO-KNOCK & 6.05 & PLANE & 4.11 \\
BUTTERFLY & 5.94 & BALL & 4.05 \\
TO-CRASH & 5.79 & DOOR & 4.0 \\
SNAKE & 5.74 & WHEELCHAIR & 4.0 \\
TO-FLY & 5.74 & CHICKEN & 3.83 \\
TABLE & 5.7 & BLANKET & 3.8 \\
PISTOL & 5.61 & CELL & 3.8 \\
EAGLE & 5.53 & DRILL & 3.8 \\
TO-CUT & 5.51 & TO-PLAY-CARDS & 3.8 \\
LAPTOP & 5.44 & BOTTLE & 3.68 \\
UMBRELLA & 5.42 & CAT & 3.61 \\
TO-JUGGLE & 5.42 & SUITCASE & 3.6 \\
CAMEL & 5.4 & LOBSTER & 3.5 \\
SPOON & 5.3 & TO-PUT-CLOTHES-ON & 3.32 \\
TO-STEAL & 5.11 & BED & 3.21 \\
TOWEL & 5.1 & RESTAURANT & 3.21 \\
BOX & 5.05 & RABBIT & 3.16 \\
\bottomrule
\end{longtable}
\subsection*{Arbitrary Signs (n = 32)}
\begin{longtable}{lc|lc}
\toprule
\textbf{Sign} & \textbf{Iconicity rating} & \textbf{Sign} & \textbf{Iconicity rating} \\
\midrule
\endhead
AMBULANCE & 3.11 & MUMMY & 2.06 \\
TO-ARGUE & 2.94 & KIWI & 2.06 \\
BEAR & 2.89 & TO-GOSSIP & 2.0 \\
TO-SHOUT & 2.79 & TO-GO-OUT & 1.89 \\
INTERPRETER & 2.74 & TOILET & 1.79 \\
DOG & 2.69 & ELECTRICITY & 1.79 \\
TO-DIE & 2.58 & PRAM & 1.74 \\
PERSON & 2.53 & DOCTOR & 1.74 \\
TO-ORDER & 2.44 & BUS & 1.74 \\
TREE & 2.26 & HORSE & 1.67 \\
TO-LAUGH & 2.26 & WATER & 1.63 \\
SOFA & 2.26 & BUILDING & 1.63 \\
ROOM & 2.22 & PUPPET & 1.53 \\
SHEEP & 2.16 & FRUIT & 1.47 \\
FIRE & 2.11 & SUGAR & 1.37 \\
TO-COOK & 2.11 & LIGHTBULB & 1.22 \\
\bottomrule
\end{longtable}

\clearpage
\twocolumn

\section{Criteria for Phonological Feature Annotation}
\label{criteria}
The following guidelines summarize the decision criteria we applied when annotating the five
phonological features of each NGT sign.  Our annotations were mainly based on the descriptions drawn from the phonology
chapters of \textit{A Grammar of Sign Language of the Netherlands (NGT)} \cite{KlompPfau2020_NGT}. We follow the general phonological descriptions in the NGT grammar but use our own simplified label set for annotation and model evaluation.
\vspace{1ex}

\textbf{Handshape:} Handshapes were coded using seven discrete labels:

\begin{itemize}\setlength\itemsep{0pt}
  \item All fingers closed to a fist
  \item All fingers extended
  \item All fingers curved or clawed
  \item One (selected) finger extended
  \item One (selected) finger curved or clawed
  \item Two or more (selected) fingers extended
  \item Two or more (selected) fingers curved or clawed
\end{itemize}

These categories are drawn from the NGT phonological inventory, but we simplify them by collapsing sub-types and by omitting features such as orientation or aperture change. However, our labels treat each sign as having a single static handshape; they therefore do not fully capture signs in which the handshape itself changes over time. For example, signs where a fist closes or opens during the articulation. For such dynamically changing signs we accepted \emph{multiple answers as correct}, so that both start and end configurations are treated as valid.
\vspace{1ex}

\textbf{Location:} Each sign was assigned to one of five major location categories:

\begin{itemize}\setlength\itemsep{0pt}
  \item Hands touching head/face
  \item Hands touching torso
  \item Hands touching arm
  \item Hands touching weak/passive hand
  \item Hands in front of the body or face (neutral space)
\end{itemize}

If a sign involved contact with multiple regions, the primary lexical target location was coded.

\vspace{1ex}

\textbf{Path Shape:} Primary path movement was classified using four labels:

\begin{itemize}\setlength\itemsep{0pt}
  \item Hold: no path or directional movement
  \item Straight: linear horizontal, vertical, or diagonal trajectory
  \item Arched: curved or semicircular trajectory
  \item Circular: full or near-full circular path
\end{itemize}

\vspace{1ex}

\textbf{Path Repetition:} Repetition of the movement was coded as:

\begin{itemize}\setlength\itemsep{0pt}
  \item Single: one primary stroke
  \item Repeated: movement is repeated

\end{itemize}
\vspace{1ex}

\textbf{Handedness:} Handedness was coded according to the two-handed typology:
\begin{itemize}\setlength\itemsep{0pt}
    \item One-handed
    \item Two-handed symmetrical: both hands share the same handshape and movement
    \item Two-handed asymmetrical: hands differ in handshape and/or movement
\end{itemize}

\section{Used Prompts}
\label{prompts}
\noindent Phonological Form Prediction Instructions:
\begin{quote}\small
\emph{Handshape?} H1=all fingers closed to a fist, H2=all fingers extended, H3=all fingers curved or clawed, H4=one (selected) finger extended, H5=one (selected) finger curved or clawed, H6= two or more (selected) fingers extended, H7=two or more(selected) fingers curved or clawed) \vspace{0.05cm}
\end{quote}
\begin{quote}\small
\emph{Location?} Major sign location? Answer with only one: L1, L2, L3, L4, L5 (L1=hands touching head/face, L2=hands touching torso, L3=hands touching arm, L4=hands touching weak/passive hand, L5=hands in front of the body or face)
\vspace{0.05cm}
\end{quote}
\begin{quote}\small
\emph{Path Shape?} Movement path shape? Answer with only one: Hold, Straight, Arched, Circular. (Hold=no path or direction, Straight=move in a straight line, Arched=move in an arched line, Circular=move in a circular path)
\vspace{0.05cm}
\end{quote}
\begin{quote}\small
\emph{Path Repetition?} Answer with only one: Single, Repeated. (Single=one movement, Repeated=multiple or repeated movements) \vspace{0.05cm} \vspace{0.05cm}
\end{quote}
\begin{quote}\small
\emph{Handedness?} Answer with only one: One-handed, Two-handed symmetrical, Two-handed asymmetrical. (One-handed=only one hand is used in the sign, Two-handed symmetrical=two hands are used but the hands move together and have the same handshape, Two-handed asymmetrical=two hands are visible, but one hand does not move and the hands have different handshapes)
\end{quote}

\noindent Transparency (open-set over 96 glosses) and Transparency$_2$ (10-choice Instructions)
\begin{quote}\small
\emph{What does this sign resemble? Look at the form and movement of the sign. Choose the most likely option from these possibilities: <OPTIONS>. Answer with only the exact word from the list.}
\end{quote}

\noindent Iconicity Rating Instructions:
\begin{quote}\small
\emph{This sign means: <MEANING>. Some signs are iconic and some are arbitrary. Find visual resemblances between the meaning and the form of the sign. How much does the sign look like ``<MEANING>’’? Answer with only one number: 1,2,3,4,5,6,7 (1=not at all, 7=exactly).}
\end{quote}

\onecolumn

\section{Iconicity Type Examples}
\label{iconicity_examples}

\begin{figure*}[h]
\centering
\newcommand{\iconpair}[3]{%
\begin{minipage}[t]{0.23\textwidth}
  \centering
  \includegraphics[width=\linewidth,height=0.08\textheight,keepaspectratio]{#1}\\[0.3em]
  \includegraphics[width=\linewidth,height=0.08\textheight,keepaspectratio]{#2}
  \vfill
  \caption*{\small #3}
\end{minipage}%
}
\iconpair{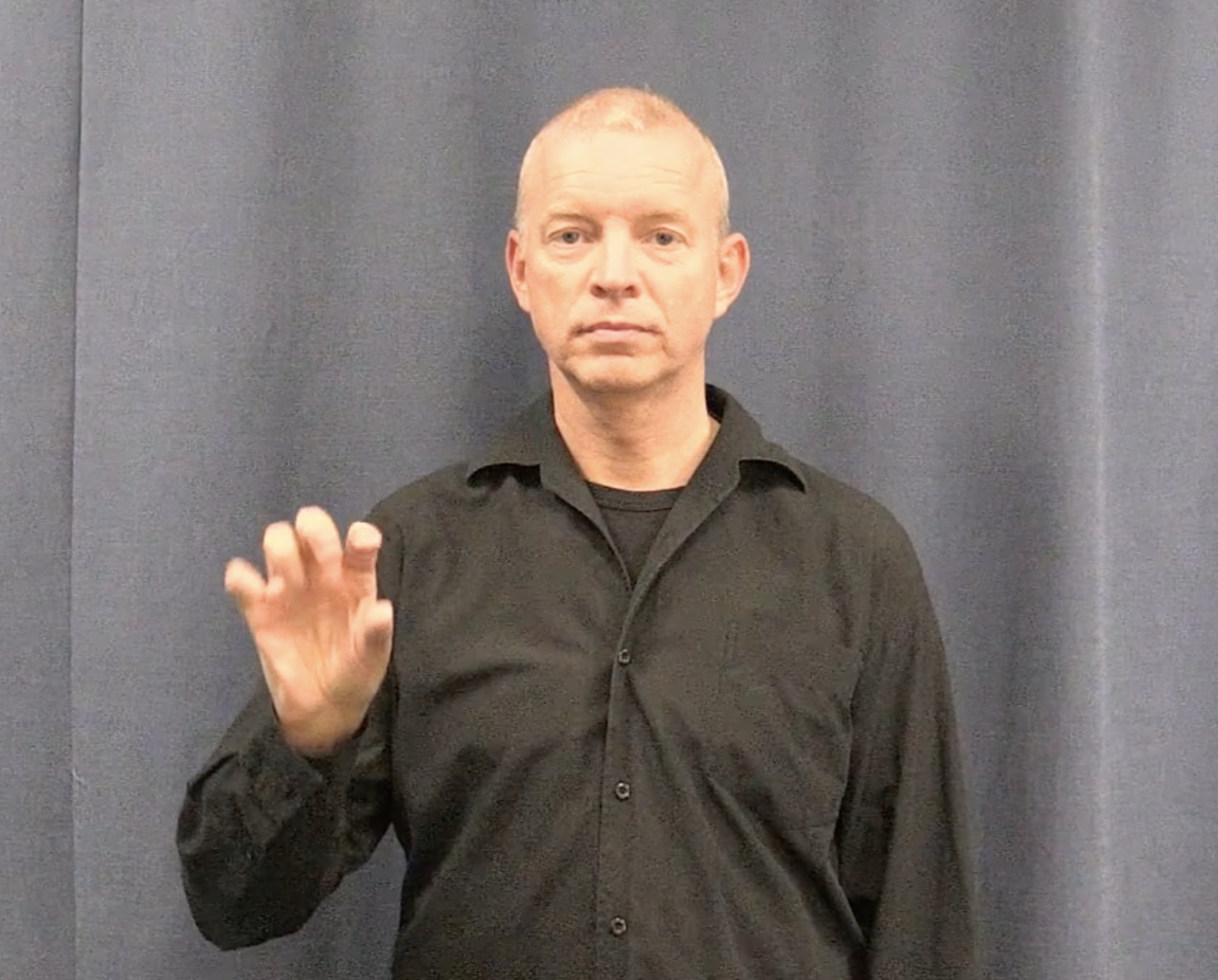}{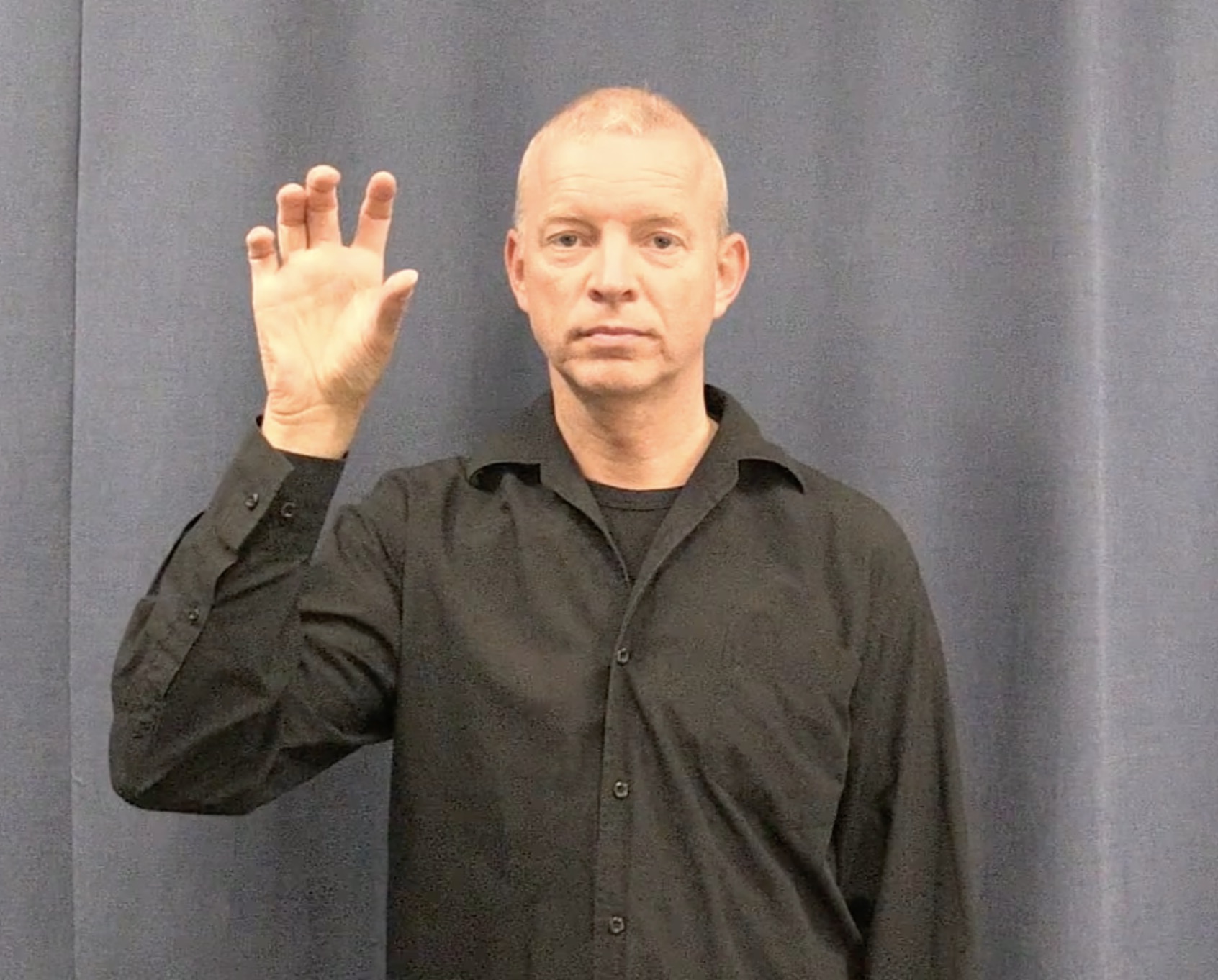}{\textbf{Combined iconic}\\\textsc{spider}: The wiggling motion of the hands conveys the spider's movement, while the curved fingers depict its legs.}\hfill
\iconpair{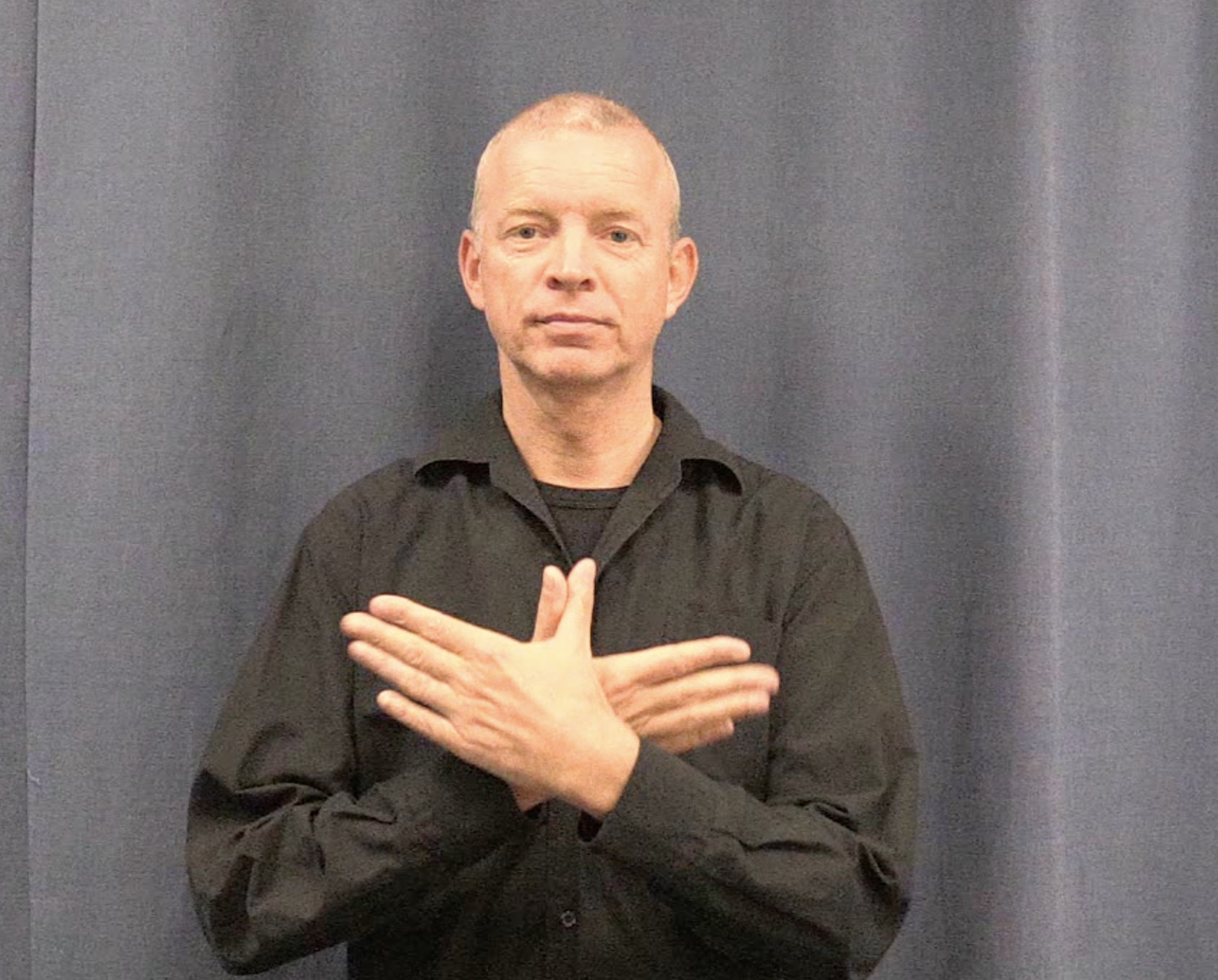}{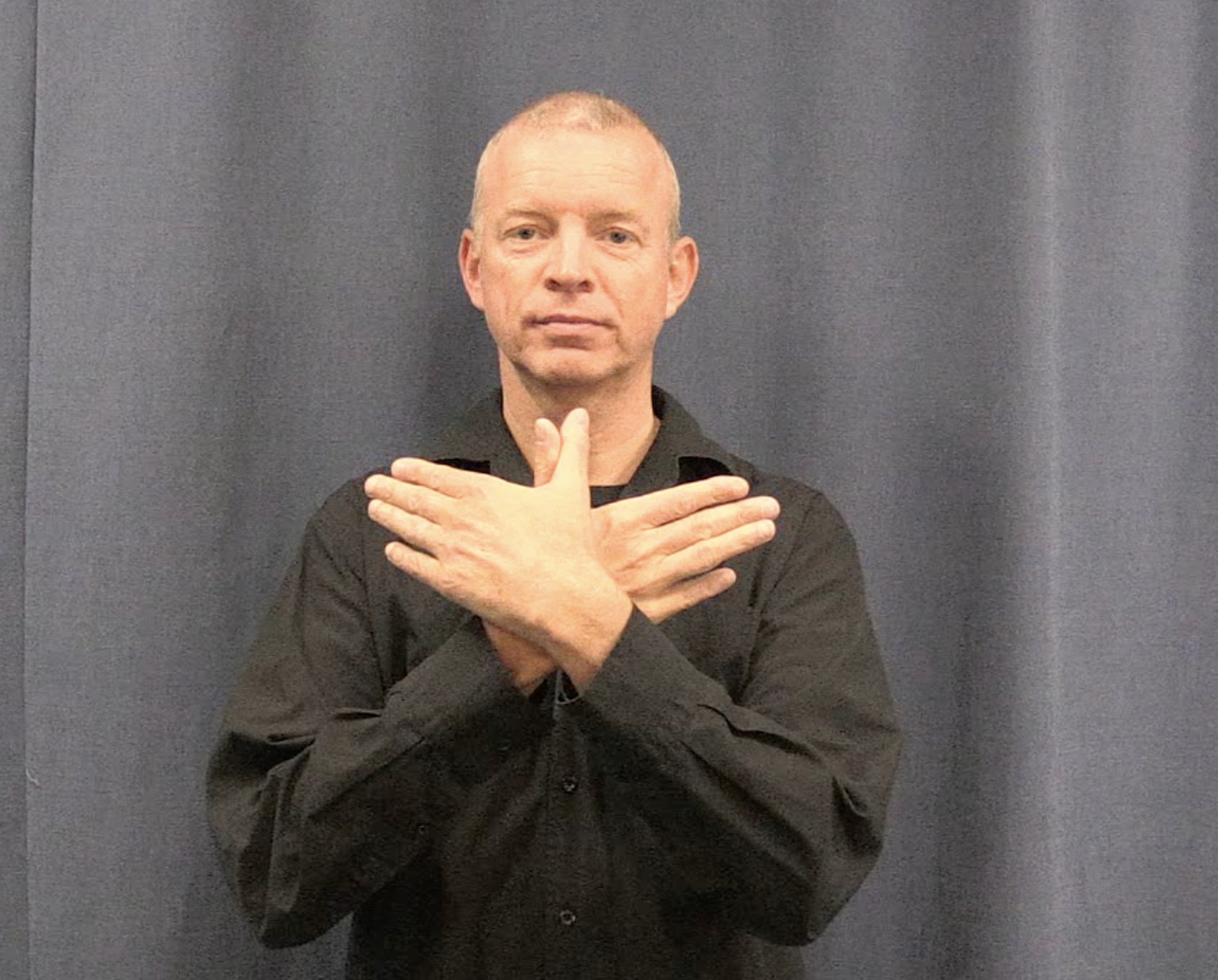}{\textbf{Object-based iconic}\\\textsc{butterfly}: two hands mirror the referent's \emph{wings}.}\hfill
\iconpair{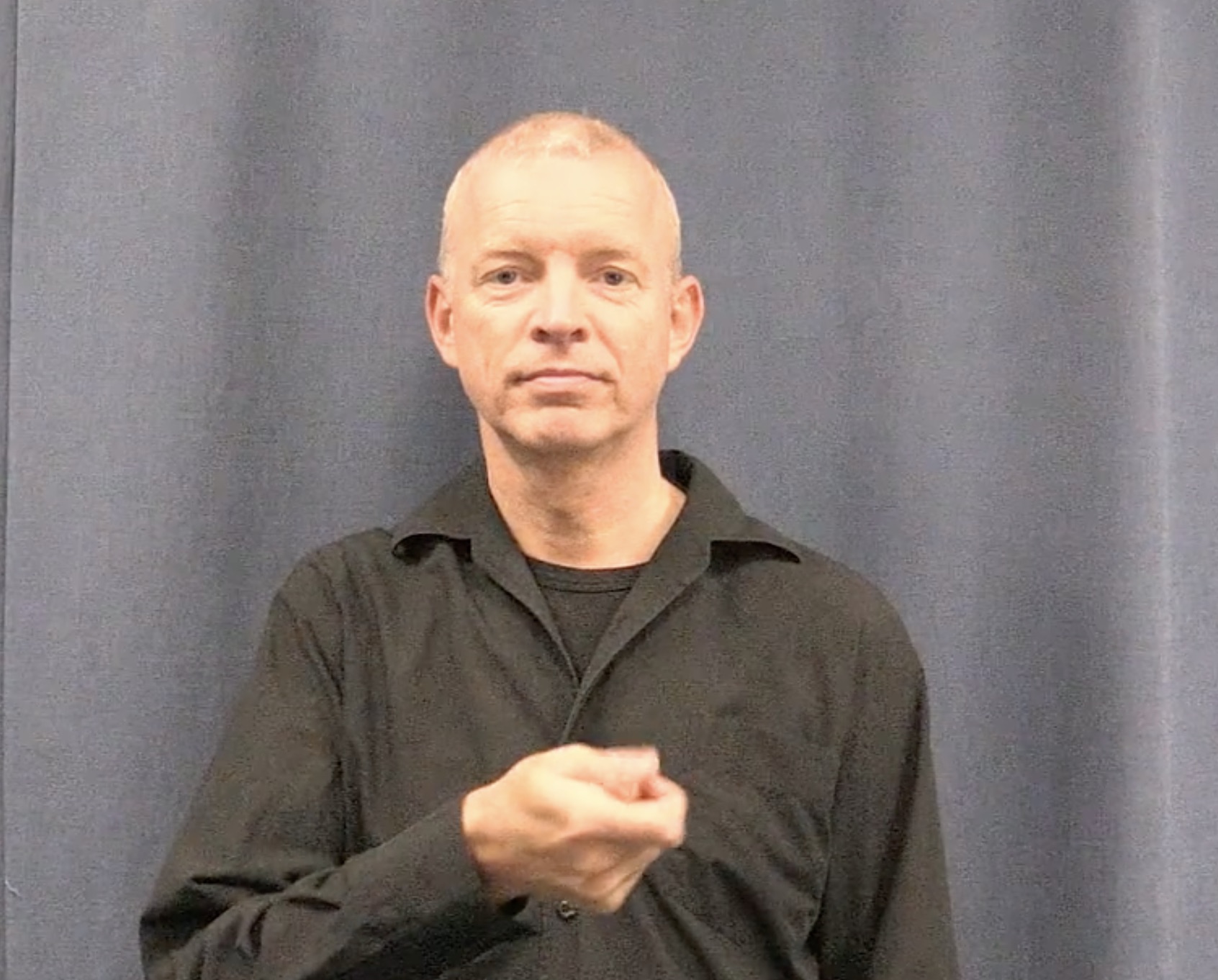}{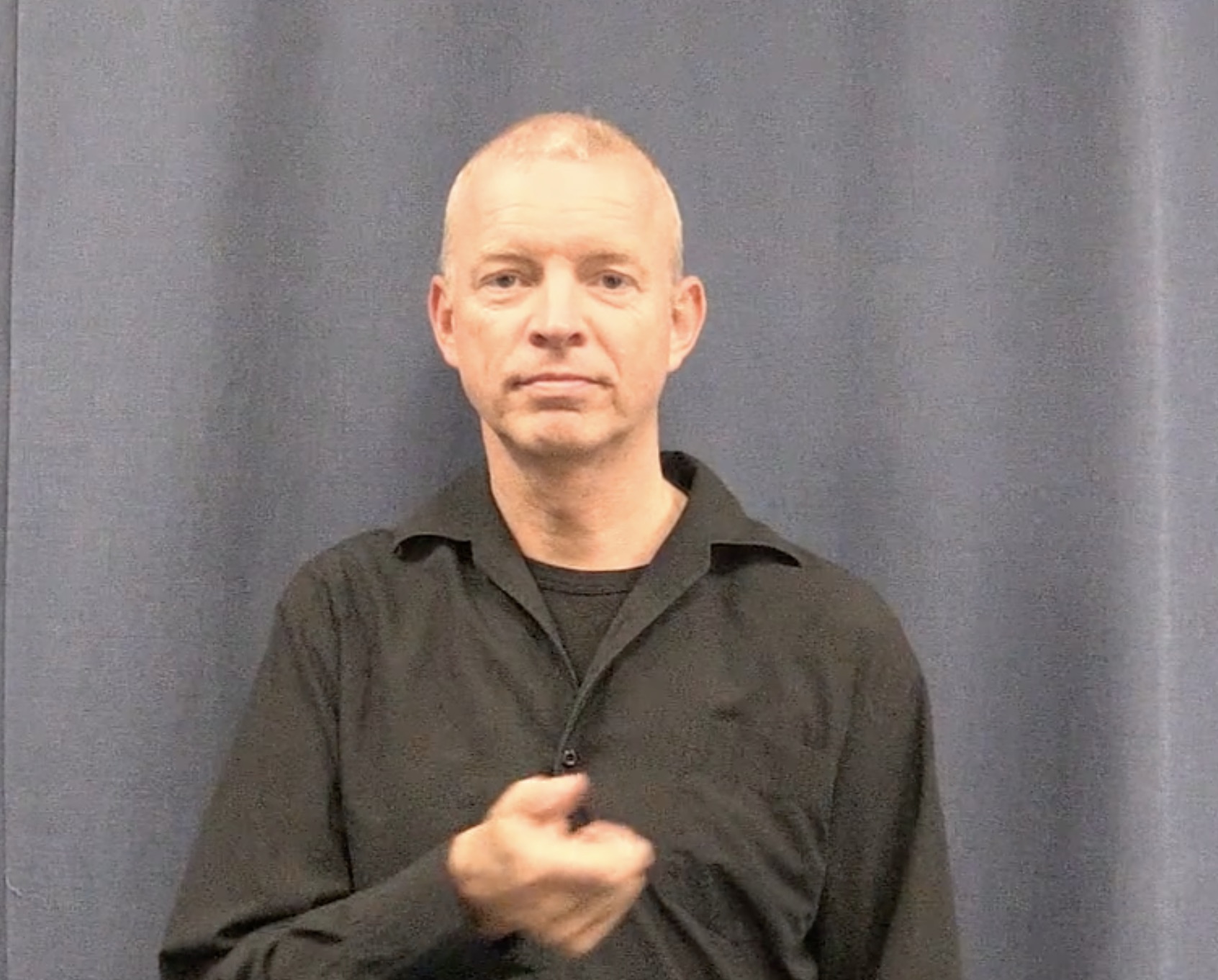}{\textbf{Action-based iconic}\\\textsc{to-sms}: thumb and fingers enact the \emph{typing/texting} action.}\hfill
\iconpair{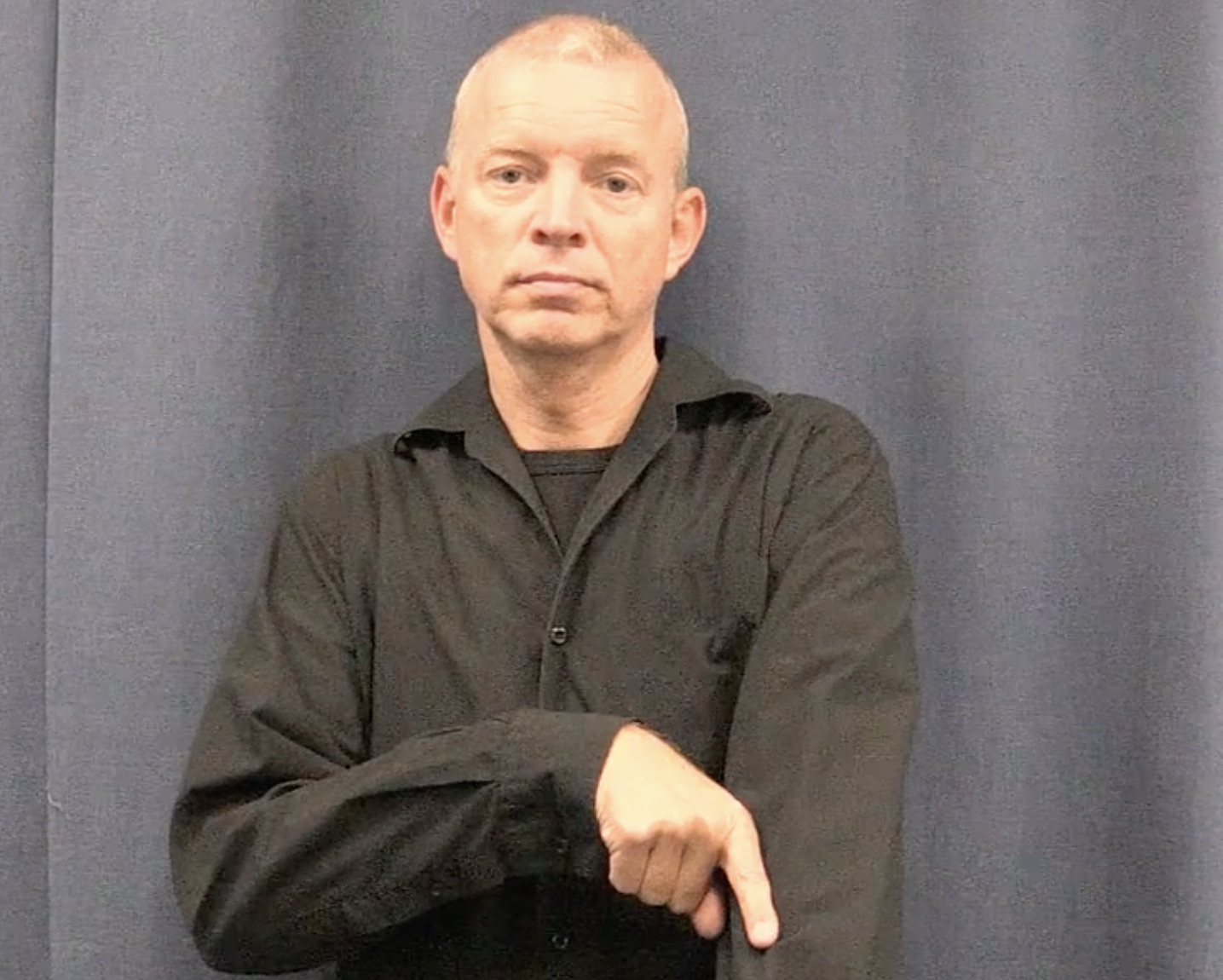}{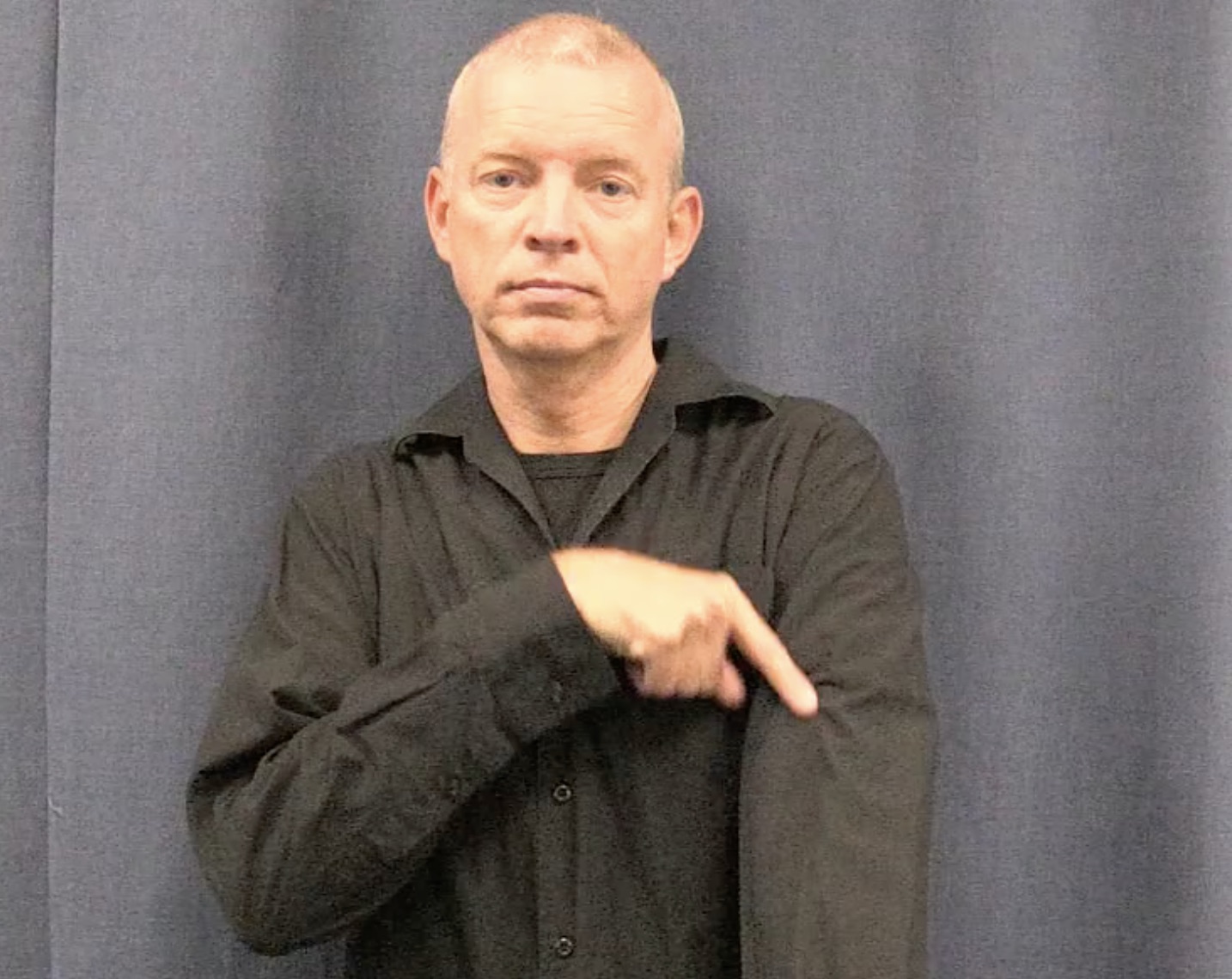}{\textbf{Arbitrary}\\\textsc{electricity}: hand configuration and path show no transparent visual resemblance; the form–meaning link is purely conventional.}

\caption{Representative frames illustrating the four iconicity categories. Each pair of frames shows how the sign fits its category.}
\end{figure*}

\section{Full Results Tables for Phonology}
\label{sect:full_results_phonology}

\begin{table*}[h]
\small
\begin{tabular}{lcccccc}
\toprule
Model & Handshape & Location & Path Shape & Path Repetition & Handedness & Mean \\
\midrule
Human baseline (hearing non-expert) & 0.698 & 0.823 & 0.677 & 0.833 & 0.938 & 0.794 \\
\midrule
\multicolumn{7}{l}{\textit{Closed-source models}}\\
Gemini-3.1-Pro              & \bfseries 0.677 & 0.844 & 0.469 & 0.719 & 0.917 & \bfseries 0.725 \\
GPT-5.4                     & 0.594 & 0.823 & \bfseries 0.490 & \bfseries 0.812 & 0.875 & 0.719 \\
Gemini-3-Flash              & 0.583 & 0.833 & 0.469 & 0.656 & 0.896 & 0.688 \\
Gemini-2.5-Pro              & \bfseries 0.677 & \bfseries 0.865 & 0.417 & 0.646 & 0.927 & 0.706 \\
GPT-5                       & 0.625 & 0.740 & 0.468 & 0.708 & \bfseries 0.948 & 0.698 \\
GPT-4o                      & 0.417 & \bfseries 0.865 & 0.365 & 0.562 & 0.917 & 0.625 \\
\midrule
\multicolumn{7}{l}{\textit{Open-source models}}\\
Qwen2.5-VL-72B              & 0.490 & 0.771 & 0.344 & 0.563 & 0.823 & 0.598 \\
MiMo-VL-7B               & 0.427 & 0.771 & 0.271 & 0.552 & 0.875 & 0.579 \\
Qwen3-VL-32B                & 0.427 & 0.594 & 0.417 & 0.531 & 0.865 & 0.567 \\
LLaVA-OV-72B          & 0.302 & 0.677 & 0.438 & 0.479 & 0.865 & 0.552 \\
Qwen2.5-VL-32B              & 0.417 & 0.719 & 0.354 & 0.563 & 0.708 & 0.552 \\
VideoLLaMA2-72B             & 0.323 & 0.729 & 0.188 & 0.563 & 0.917 & 0.544 \\
Gemma-3-27B                 & 0.333 & 0.781 & 0.302 & 0.552 & 0.708 & 0.535 \\
Qwen3-VL-4B                 & 0.333 & 0.448 & 0.396 & 0.542 & 0.802 & 0.504 \\
Qwen2.5-VL-7B               & 0.427 & 0.385 & 0.188 & 0.552 & 0.531 & 0.417 \\
LLaVA-OV-7B           & 0.167 & 0.167 & 0.313 & 0.604 & 0.802 & 0.411 \\
VideoLLaMA2-7B              & 0.083 & 0.135 & 0.188 & 0.000 & 0.198 & 0.121 \\
\midrule
Random baseline              & 0.143 & 0.200 & 0.250 & 0.500 & 0.333 & 0.285 \\
\bottomrule
\end{tabular}
\caption{Phonological form prediction accuracy by model and phonological subtasks, with random and human baselines, and mean accuracy across all subtasks.}
\label{tab:accuracy_by_model_handshape_sorted_mean}
\end{table*}


\newpage
\section{Additional Figures}
\label{sect:additional_results}
\begin{figure}[h]
    \centering
\includegraphics[width=0.5\linewidth]{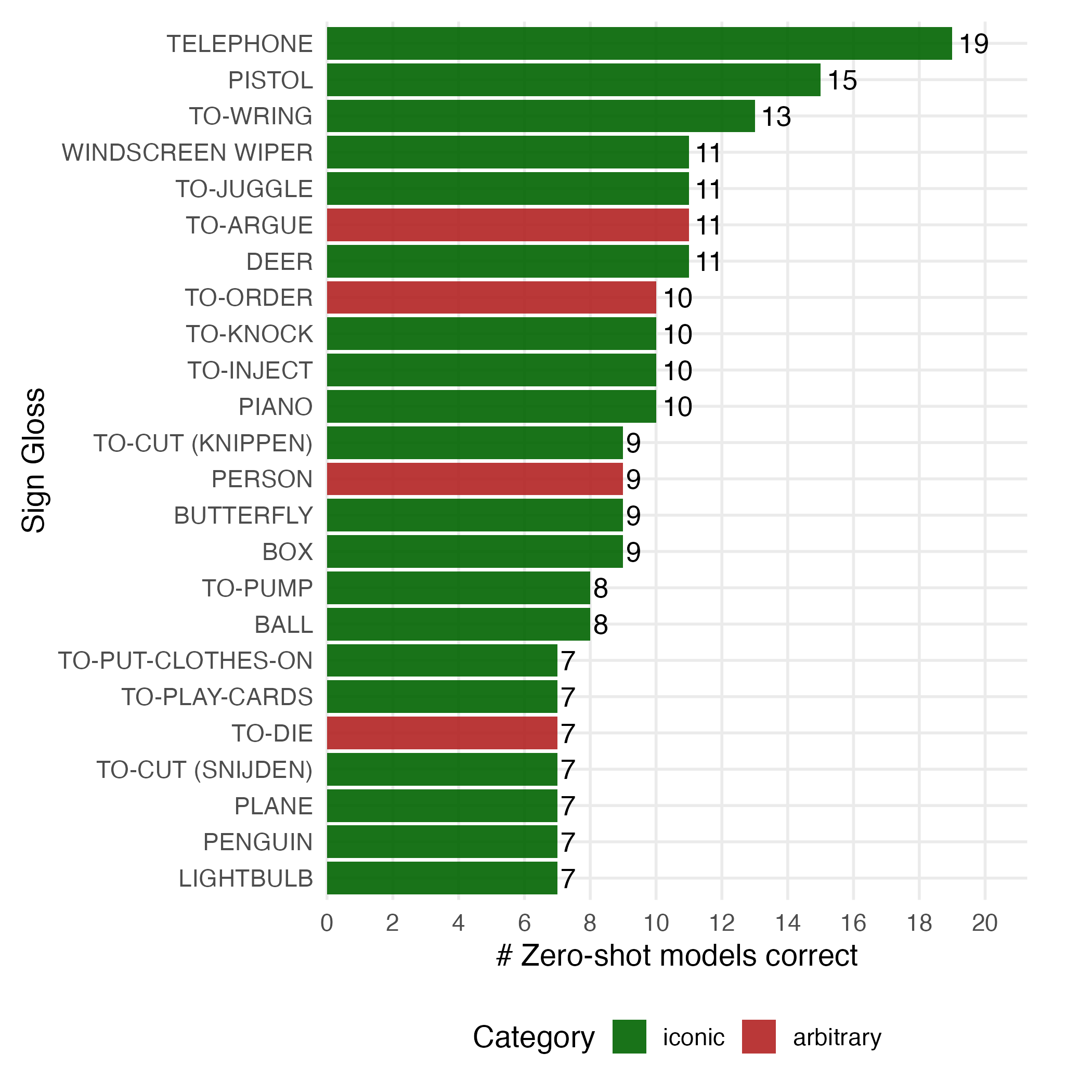}
    \caption{Correctly guessed signs from sign video only ($\geq 7$ VLMs) in the Transparency$_2$ Task. Bars in green mark iconic signs; red bars mark arbitrary signs that were nonetheless guessed by multiple VLMs.}
    \label{fig:transparency_common_glosses}
\end{figure}

\begin{figure}[h]
    \centering
    \includegraphics[width=0.5\linewidth]{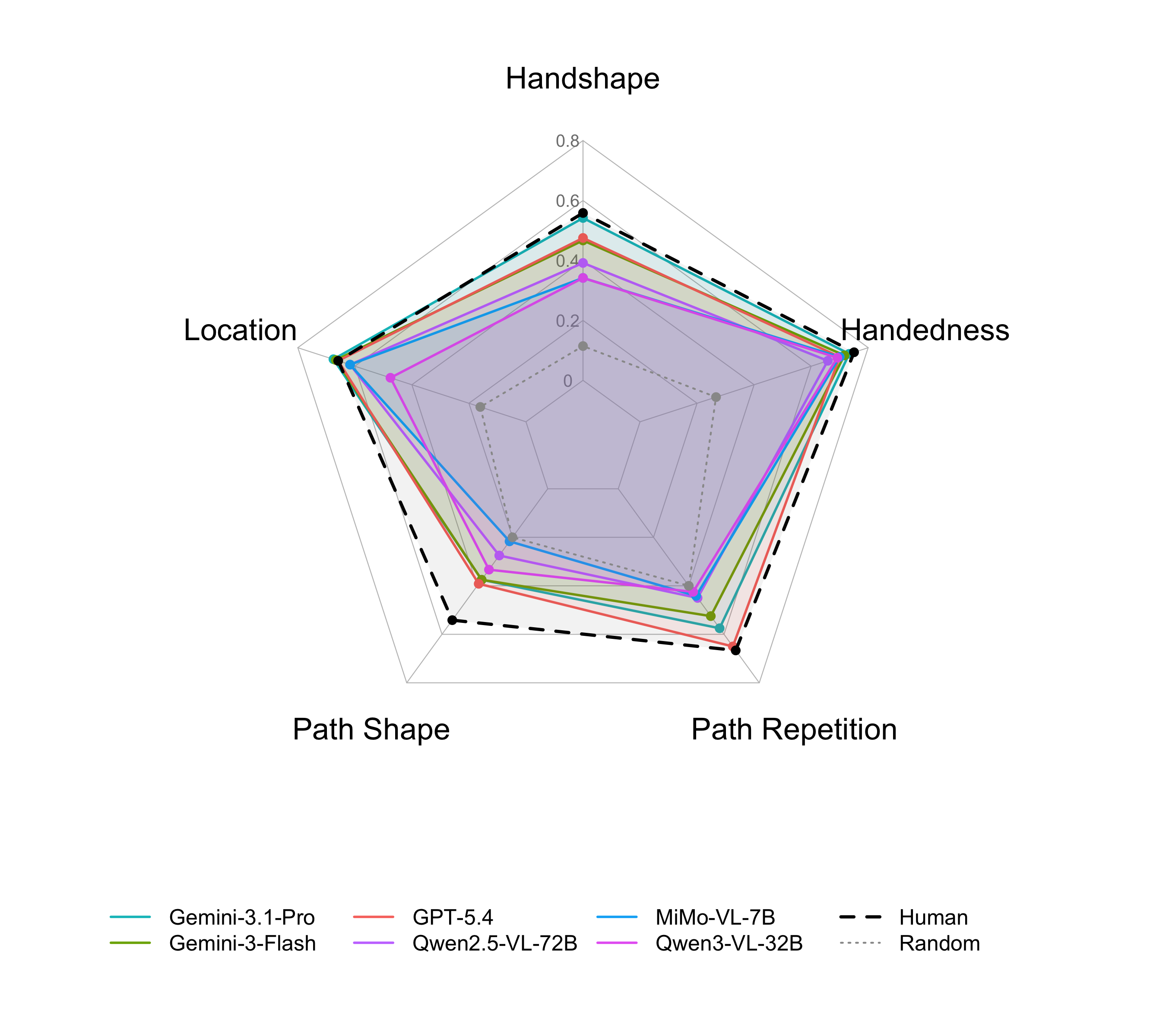}
    \caption{Zero-shot accuracy of the top six performing models (3 closed: Gemini~3.1~Pro, Gemini~3~Flash, GPT-5.4; 3 open: Qwen2.5-VL-72B, MiMo-VL-7B, Qwen3-VL-32B) across five phonological features, averaged over 96 signs. Solid colored lines are models; the dashed black line is the human (deaf-native) baseline, and the grey dotted line is the random baseline.}
    \label{fig:form_features_summary}
\end{figure}

\clearpage
\section{Qualitative Error Analysis}
\label{sect:error_analysis}

To complement the quantitative results, we conducted a qualitative inspection of signs with the highest discrepancy between human and the best-performing open-source VLM ratings. We identified three recurring failure modes which we illustrate with representative examples in Table~\ref{tab:err_examples}. The categories highlight \emph{what} models fail to perceive in signed forms, not only \emph{how often} they fail.

\begin{table}[H]
\small
\centering
\setlength{\tabcolsep}{3pt}
\renewcommand{\arraystretch}{1.1}
\begin{tabular}{p{0.16\linewidth}p{0.78\linewidth}}
\toprule
\textbf{Missing Agent} &
NGT signs often depict an agent manipulating an object through transitive actions. VLMs appear to be trained predominantly on static, isolated objects and fail to map the holding or manipulating action to the referent noun. \emph{Examples:} \textsc{baby} (cradling motion); \textsc{calculator} (finger pressing keys on held object); \textsc{towel} (wiping motion); \textsc{car} (gripping and turning steering wheel). \\
\addlinespace
\textbf{Static Bias} &
Models rate signs with static handshapes higher than those with dynamic motions, suggesting better alignment with static visual contours than dynamic event simulations. VLMs privilege shape-matching over motion-based semantic grounding, inverting the human preference for action-based iconicity. \emph{Example:} \textsc{to-fly} (static airplane classifier handshape rated higher than dynamic \textsc{bird} with flapping wings). \\
\addlinespace
\textbf{Gloss Sensitivity} &
Visual forms that are historically specific, culturally bound, or incongruent with contemporary web-scraped training data cause systematic failures. \emph{Examples:} \textsc{to-sms} (single-handed typing on old keypad phone; web data shows two-thumbed smartphone typing); \textsc{zimmer/rollator} (walking frame uncommon in training data); \textsc{to-steal} (pickpocketing with peripheral twisting motion; web images show generic theft scenes). \\
\bottomrule
\end{tabular}
\caption{Three recurring qualitative failure modes observed in zero-shot VLM responses on signs with the largest model--human discrepancy. The \emph{Missing Agent} and \emph{Static Bias} modes correspond to the under-rating of action-based iconicity in Figure~\ref{fig:iconicity_type}; \emph{Gloss Sensitivity} reflects limitations of web-scraped training data.}
\label{tab:err_examples}
\end{table}

\end{document}